\documentclass[a4paper,fleqn]{cas-sc}

\usepackage[numbers]{natbib}
\usepackage{arydshln} 
\def\tsc#1{\csdef{#1}{\textsc{\lowercase{#1}}\xspace}}
\tsc{WGM}
\tsc{QE}
\tsc{EP}
\tsc{PMS}
\tsc{BEC}
\tsc{DE}

\usepackage{hyperref}
\hypersetup{hidelinks,
	colorlinks=true,
	allcolors=blue,
	pdfstartview=Fit,
	breaklinks=true}
\usepackage{amsmath}
\usepackage{amssymb}
\usepackage{multirow}
\usepackage{adjustbox}
\usepackage{makecell}
\usepackage{subcaption}
\usepackage{overpic}  
\usepackage{booktabs}
\usepackage{float}
\usepackage{algorithm}
\usepackage{algpseudocode}
\usepackage{graphicx}  

\usepackage{lineno}

\begin{document}
\let\WriteBookmarks\relax
\def\floatpagepagefraction{1}
\def\textpagefraction{.001}

% Short title
\shorttitle{MSSDF}

% Short author
\shortauthors{Tong Wang et~al.}

% Main title of the paper
\title [mode = title]{MSSDF: Modality-Shared Self-supervised Distillation for High-Resolution Multi-modal Remote Sensing Image Learning}                      

\author[1]{Tong Wang}[type=editor,
                        orcid=0000-0002-3915-0175,style=chinese]

\fnmark[1]
\credit{Conceptualization of this study, Methodology, Software, Data curation, Writing - Original draft preparation}

\author[1]{Guanzhou Chen}[style=chinese]
\fnmark[1]
\cormark[1]
\ead{cgz@whu.edu.cn}
\credit{Data curation, Writing - Original draft preparation, Funding acquisition}
\author[1]{Xiaodong Zhang}[style=chinese]
\cormark[1]
\ead{zxdlmars@whu.edu.cn}
\credit{Supervision, Project administration, Funding acquisition}

\author[2]{Chenxi Liu}[style=chinese]
\credit{Writing - Review \& Editing, Visualization}
\author[1]{Jiaqi Wang}[style=chinese]  
\credit{Writing - Review \& Editing, Resources}
\author[1]{Xiaoliang Tan}[style=chinese]
\credit{Writing - Review \& Editing, Validation}
\author[1]{Wenchao Guo}[style=chinese]
\credit{Writing - Review \& Editing}
\author[1]{Qingyuan Yang}[style=chinese]
\credit{Writing - Review \& Editing}
\author[1]{Kaiqi Zhang}[style=chinese]
\credit{Writing - Review \& Editing}

\affiliation[1]{organization={State Key Laboratory of Information Engineering in Surveying, Mapping and Remote Sensing, Wuhan University},
            addressline={No.129, Luoyu Road}, 
            city={Wuhan},
            postcode={420079}, 
            state={Hubei},
            country={China}}
\affiliation[2]{organization={Electronic Information School, Wuhan University},
    addressline={299 Bayi Road, Wuchang District}, 
    city={Wuhan},
       postcode={430072}, 
    country={China}}

\cortext[cor1]{Corresponding authors: cgz@whu.edu.cn (G. Chen), zxdlmars@whu.edu.cn (X. Zhang).}

\fntext[fn1]{These authors contributed equally.}

% Here goes the abstract
\begin{abstract}
Remote sensing image interpretation plays a critical role in environmental monitoring, urban planning, and disaster assessment. However, acquiring high-quality labeled data is often costly and time-consuming. To address this challenge, we proposes a multi-modal self-supervised learning framework that  leverages high-resolution RGB images, multi-spectral data, and digital surface models (DSM) for pre-training. By designing an information-aware adaptive masking strategy, cross-modal masking mechanism, and multi-task self-supervised objectives, the framework effectively captures both the correlations across different modalities and the unique feature structures within each modality.  We evaluated the proposed method on multiple downstream tasks, covering typical remote sensing applications such as scene classification, semantic segmentation, change detection, object detection, and depth estimation. Experiments are conducted on 15 remote sensing datasets, encompassing 26 tasks. The results demonstrate that the proposed method outperforms existing pretraining approaches in most tasks. Specifically, on the Potsdam and Vaihingen semantic segmentation tasks, our method achieved mIoU scores of 78.30\% and 76.50\%, with only 50\% train-set. For the US3D depth estimation task, the RMSE error is reduced to 0.182, and for the binary change detection task in SECOND dataset, our method achieved mIoU scores of 47.51\%,  surpassing the second CS-MAE by 3 percentage points. In addition, we construct a high-resolution multi-modal remote sensing image dataset named HR-Paris, which contains 640,000 DOM-DSM image pairs with a spatial resolution of 0.05 meters, providing a new high-quality dataset for multi-modal remote sensing research. Ablation studies further validate the effectiveness of multi-modal fusion and the contribution of each self-supervised task component to the final performance. Our pretrain code, checkpoints, and HR-Pairs dataset can be found in \url{https://github.com/CVEO/MSSDF}.
\end{abstract}

% Use if graphical abstract is present
% \begin{graphicalabstract}
% \includegraphics{figs/grabs.pdf}
% \end{graphicalabstract}

% Research highlights
\begin{highlights}

    \item Proposed a modality-shared self-supervised distillation framework for pertraining.

   \item Designed an information-aware strategy to keep semantic important content.

    \item Introduced a cross-modal mask method to enhance inter-modality correlation learning.

    \item Formulated a multi-task objective for better pretrain task.
    
    \item Built a 640K 5cm-resolution multi-modal remote sensing dataset.

\end{highlights}

% Keywords
% Each keyword is seperated by \sep
\begin{keywords}
Multimodal Data Fusion \sep Remote Sensing \sep Lightweight Network \sep Semantic Segmentation \sep Digital Surface Model
\end{keywords}

\maketitle

% \linenumbers

%% main text
\section{Introduction}
\label{sec:intro}

In recent years, the rapid development of remote sensing and drone technologies has enabled the acquisition of ultra-high-resolution (UHR) imagery at unprecedented scales~\cite{luo2025deep,christie2018functional,long2021creating}. These data provide rich spatial, spectral, and temporal information, offering fundamental support for a wide range of applications such as urban planning, land use monitoring, and environmental assessment~\cite{lu2025vision}. For instance, centimeter-level 3D reconstructions available through \textit{Google Earth} have reached Petabyte-scale datasets~\cite{googleearth_2024,gibson2017io}, indicating that high-resolution remote sensing imagery is becoming a vital resource for surface observation.

\begin{figure}[htb]
    \centering
    \includegraphics[width=0.7\textwidth]{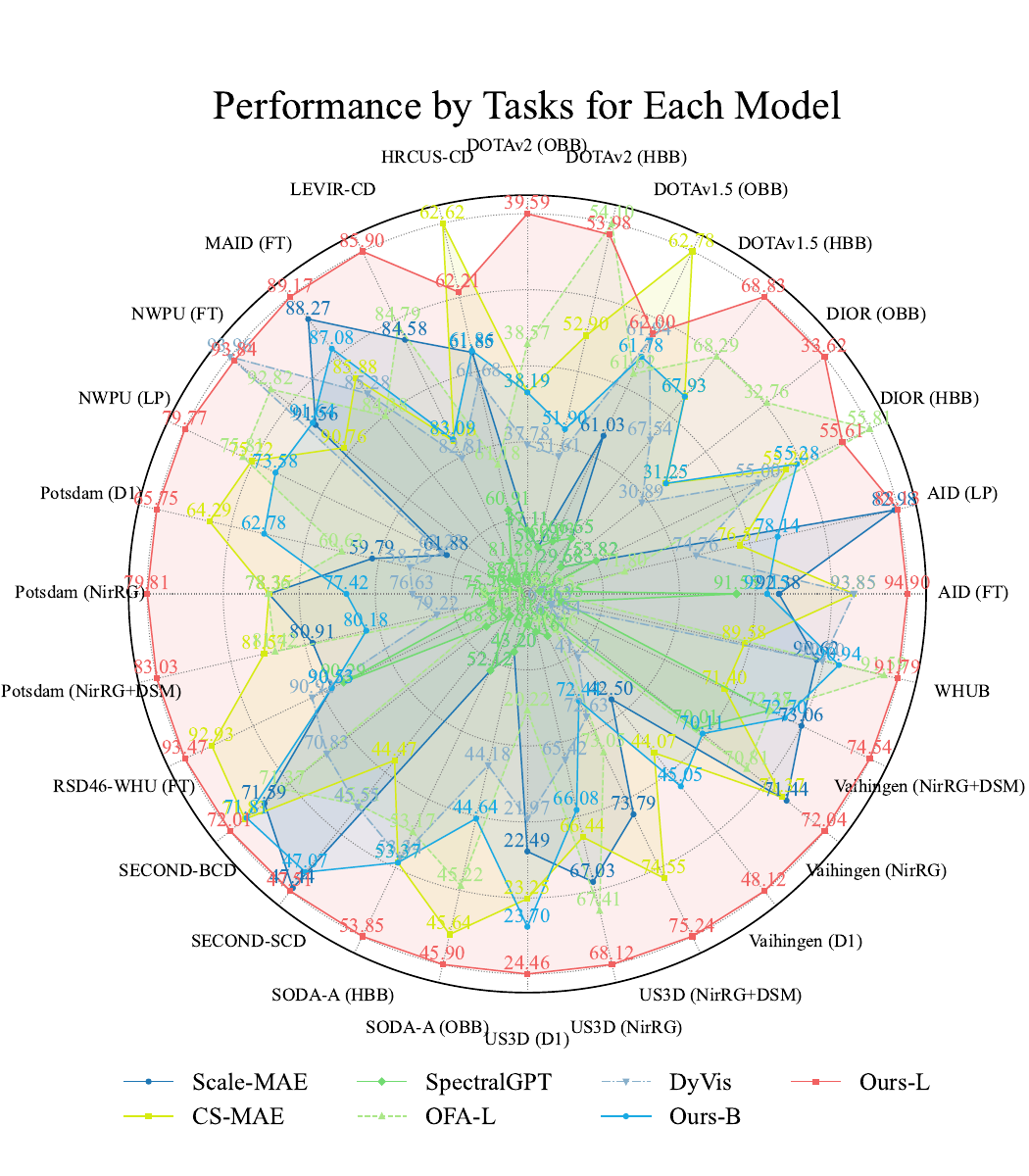}
    \caption{Our model has achieved state-of-the-art performance on the most of tasks across 26 tasks evaluated on 15 datasets.}
    \label{fig:radar}
\end{figure}

With the increasing availability of multi-modal remote sensing data — including optical,  Synthetic Aperture Radar (SAR), LiDAR, and Digital Surface Models (DSM)~\cite{zhang2025cdf,wang2025lmfnet}  — there is growing interest in leveraging complementary information across modalities to improve scene understanding and downstream applications~\cite{christie2018functional,leite2025multimodal}. However, existing methods often rely heavily on labeled datasets and struggle to capture cross-modal correlations in a unified feature space~\cite{wang2025uncertainty}. Moreover, current self-supervised frameworks either ignore modality-specific characteristics or fail to preserve critical semantic information under high-resolution settings, limiting their practical effectiveness.

Despite the emergence of large-scale pretraining datasets such as MillionAID~\cite{long2021creating}, SeCo~\cite{manas2021seasonal}, SSL4EO-S12~\cite{wang2023ssl4eo}, and BigEarthNet~\cite{sumbul2021bigearthnet}, most suffer from two key limitations: \textbf{(1) limited spatial resolution}, typically at the meter level or coarser, which restricts detailed analysis of UHR scenes; and \textbf{(2) lack of dense pixel-level alignment} between heterogeneous modalities (e.g., DOM and DSM), hindering effective joint modeling and reconstruction. These issues are particularly pronounced in complex urban environments where fine-grained structural details are essential.

\begin{figure}[htb]
    \centering
    \includegraphics[width=0.5\textwidth]{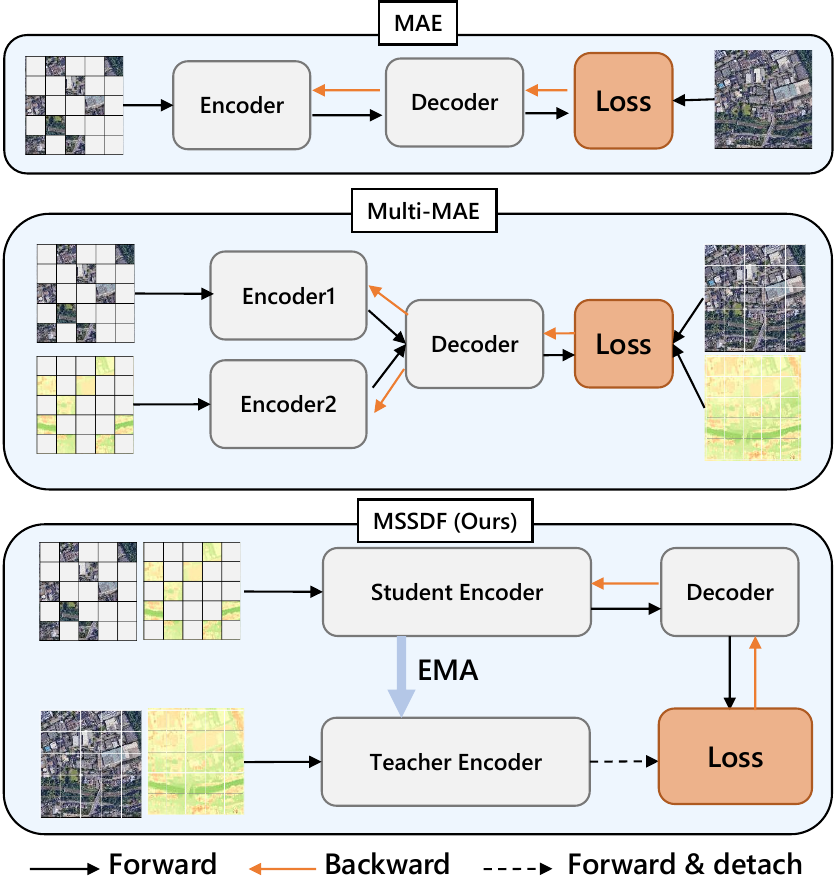}
    \caption{Architectures of Masked Image Modeling (MIM) Approaches for Multi-Modal Remote Sensing Data }
    \label{fig:diff_model}
\end{figure}

Meanwhile, recent progress in masked image modeling (MIM)~\cite{he2022masked} has shown promising results in natural image domains, and has begun to influence remote sensing research. Methods such as ViT-based Mask Autoencoder (MAE) variants, FG-MAE~\cite{wang2024feature}, and SatMAE++~\cite{bastani2023satlaspretrain} have demonstrated strong transferability on RGB or multispectral imagery. However, these approaches seldom address the unique challenges posed by UHR multi-modal data, especially when dealing with spectral-spatial heterogeneity and pixel-aligned DOM-DSM pairs. Figure~\ref{fig:diff_model} illustrates how recent MIM architectures aim to tackle these challenges through different strategies for encoding and reconstructing multi-modal inputs.

To this end, in this study, we propose a novel framework called the Modality-Shared Self-supervised Distillation Framework (MSSDF) , designed specifically for high-resolution multi-modal remote sensing image analysis. Our approach integrates contrastive learning, knowledge distillation, and adaptive masking strategies to learn robust, discriminative, and consistent representations across different modalities. We further introduce an Information-Aware Adaptive Masking Strategy , which preserves semantically important regions while enabling effective cross-modal alignment. Additionally, we formulate a Multi-Task Self-Supervised Training Objective , combining reconstruction loss, contrastive alignment loss, feature decorrelation loss, and auxiliary classification loss, promoting comprehensive feature learning.

To facilitate future research and benchmarking, we also construct a new dataset named HR-Paris , which contains over 640,000 centimeter-level pixel-aligned DOM--DSM image pairs derived from drone oblique photography and Google Earth 3D Tiles. The dataset covers multiple countries and regions across the globe, and provides a realistic and diverse foundation for evaluating high-resolution multi-modal representation learning methods.

The key innovations of our work are summarized as follows:

\begin{itemize}
    \item \textbf{Modality-Shared Self-supervised Distillation Framework (MSSDF)}: Proposed a multi-modal feature learning framework based on a student-teacher architecture, contrastive learning, and knowledge distillation.
    \item \textbf{Modal-Aware Masking Strategy}: Designed an information-aware adaptive masking strategy to preserve more valuable content, and introduced a cross-modal masking mechanism to facilitate the model's learning of cross-modal correlations.
    \item \textbf{Multi-Task Self-Supervised Training Objective}: Introduced a multi-task self-supervised training objective, including reconstruction loss, contrastive alignment loss, feature decorrelation loss, and auxiliary classification loss, which helps to learn rich, consistent, and discriminative multi-modal feature representations. 
    \item \textbf{High-Resolution Multi-Modal Remote Sensing Dataset (HR-Paris)}: Constructed a high-resolution multi-modal remote sensing image dataset, which covers a variety of terrain and environmental conditions, providing a high-quality dataset for relevant research
\end{itemize}

We evaluate our methods on a diverse set of 15 datasets, as shown in Table~\ref{tab:downstream}. As shown in Figure~\ref{fig:radar}, Our model has achieved state-of-the-art performance on the majority of tasks across 26 tasks evaluated on 15 datasets. Extensive experiments demonstrate that our framework achieves superior performance on various downstream tasks, including semantic segmentation, change detection, and depth estimation, outperforming state-of-the-art methods in both accuracy and generalization. 

The following of this paper is as follows: Section~\ref{sec:relwork} introduces the relevant research progress made by previous scholars. Section~\ref{sec:hrdata} presents the methodology for constructing the HR-Pairs dataset. The subsequent  Section~\ref{sec:method} outlines our pre-training framework. The experimental Section~\ref{sec:exp} details the experimental setup, results of downstream tasks, comparisons with various pre-training methods, and ablation studies. Finally, Section~\ref{sec:conclusion} provides a summary of the work presented in this paper.

\section{Related Work}
\label{sec:relwork}
With the rapid advancement of remote sensing technologies, self-supervised pretraining has emerged as a promising approach for learning rich and transferable feature representations from large-scale unlabeled data~\cite{luo2025deep, an2024pretrain}. In recent years, significant progress has been made in three key directions: \textbf{(1)} the development of large-scale remote sensing datasets with diverse modalities and spatial resolutions~\cite{dong2024generative}; \textbf{(2)} the adaptation of masked image modeling techniques from natural vision domains to remote sensing applications~\cite{bastani2023satlaspretrain}; and \textbf{(3)} the integration of multi-modal fusion strategies with self-supervised learning frameworks~\cite{leite2025multimodal}. Additionally, mechanisms such as exponential moving average (EMA) and knowledge distillation~\cite{chen2018training,himeur2024applications,wang2024knowledge} have proven effective in stabilizing training and improving representation quality. This section provides a comprehensive review of these developments, highlighting their strengths and limitations, and positioning our proposed method within this evolving landscape. 

\subsection{Development of Remote Sensing Pretrain Datasets}

\begin{table*}[h]
\centering
\small
\caption{Overview of Remote Sensing Pretrain Datasets}
\label{tab:dataset}
\resizebox{\textwidth}{!}{% 缩放到页面宽度
\begin{tabular}{ccccc}
\hline
\textbf{Dataset Name} & \textbf{\# of Images} & \textbf{Res. (m)} & \textbf{Data Modalities} & \textbf{Data Source} \\
\hline
MillionAID~\cite{long2021creating} & 1,000,848 & 0.5-153 & RGB & Google Earth \\
SatlasPretrain~\cite{bastani2023satlaspretrain} & 856,000 & 0.5-2.10 & RGB, Multispectral & NAIP, Sentinel-2 \\
SeCo~\cite{manas2021seasonal} & 1,000,000 & 10 & Multispectral & Sentinel-2 \\
fMoW~\cite{christie2018functional} & 1,047,691 & - & Multispectral (4/8 bands) & Digital Globe \\
SEN12MS~\cite{schmitt2019sen12ms} & 180,662 & 10 & SAR-Multispectral & Sentinel-1, Sentinel-2 \\
BigEarthNet-MM~\cite{sumbul2021bigearthnet} & 1,180,652 & 10-60 & SAR-Multispectral & Sentinel-1, Sentinel-2 \\
SSL4EO-S12~\cite{wang2023ssl4eo} & 3,012,948 & 10-60 & SAR-Multispectral & Sentinel-1, Sentinel-2 \\
SSL4EO-L~\cite{stewart2023ssl4eo} & 5,000,000 & 30 & Multispectral & Landsat 4-9 Series \\
\hline
\textbf{HR-Pairs(Ours)} & 640,000 & \textbf{0.05} & DOM (RGB)-DSM & drone, Google Earth 3dtiles \\
\hline
\end{tabular}
}
\end{table*}

In recent years, numerous large-scale unsupervised or weakly supervised pretraining datasets have emerged in the remote sensing community, aiming to extract rich semantic representations through self-supervised strategies. Examples include MillionAID, SatlasPretrain, SeCo, fMoW, SEN12MS, BigEarthNet-MM, SSL4EO-S12, and SSL4EO-L, which encompass diverse modalities such as visible, multispectral, and SAR data, covering extensive geographic regions with various sensors (see Table~\ref{tab:dataset}).

While these datasets demonstrate strong generalization in remote sensing tasks, they generally suffer from two limitations: first, limited spatial resolution, which is insufficient for UHR scene analysis; second, a lack of dense pixel-level modality alignment, restricting effective joint modeling of heterogeneous data such as DOM and DSM. To fill this gap, we introduce the HR-Pairs dataset, which provides centimeter-level pixel-aligned RGB-DOM and DSM image pairs, thus enabling detailed cross-modal analysis at a fine spatial scale.

\begin{table*}[htb]
\centering
\caption{Summary of Representative Remote Sensing Self-Supervised Learning Models}
\label{tab:rs_models}
\resizebox{\textwidth}{!}{
\begin{tabular}{c|ccccccc}
\hline
 & \textbf{Scale-MAE} & \textbf{CS-MAE} & \textbf{DMC-R} & \textbf{DMC-S} & \textbf{TOV} & \textbf{GeRSP} & \textbf{PIS}  \\
\hline
\textbf{Pub. Year} & 2023 & 2023 & 2023 & 2023 & 2023 & 2024 & 2024  \\
\textbf{Backbone} & ViT-Large & ViT-Large & ResNet50 & Swin-Large & ResNet50 & ResNet50 & ResNet50 \\
\textbf{Dataset} & FMoW & FMoW & SeCo-100K & SeCo-100K & TOV-RS & Million-AID & SSL4EO-S12 \\
\textbf{Paper} & \cite{reed2023scale} & \cite{tang2023cross} & \cite{wanyan2023dino} & \cite{wanyan2023dino} & \cite{tao2023tov} & \cite{huang2024generic} & \cite{an2024pretrain} \\
\hline
& \textbf{SMLFR} & \textbf{SpectralGPT} & \textbf{FG-MAE}  & \textbf{SatMAE++} & \textbf{OFA-L} & \textbf{DyVis} & \textbf{MSSDF (Ours)} \\
\hline
\textbf{Pub. Year}  & 2024 & 2024 & 2024 & 2024 & 2024 & 2025 & 2025 \\
\textbf{Backbone} & ConvNext Large & ViT-Base & ViT-Base/Large  & ViT-Large & ViT-Large & DynamicVis &  ViT-Base/Large\\
\textbf{Dataset}& GeoSense & \makecell{FMoW \\ BigEarthNet}  & SSL4EO-S12 & FMoW-RGB & Various & FWoW-RGB &  \makecell{FMoW \\ HR-Pairs}\\
\textbf{Paper}   & \cite{dong2024generative} & \cite{hong2024spectralgpt} & \cite{wang2024feature}  & \cite{noman2024rethinking} & \cite{xiong2024one} & \cite{chen2025dynamicvis} & -\\
\hline
\end{tabular}
}
\end{table*}

\subsection{Masked Image Modeling}

Masked Image Modeling~\cite{he2022masked} (MIM) has achieved remarkable progress in natural image self-supervised learning and has been gradually adapted for remote sensing applications~\cite{reed2023scale}. As shown in Table~\ref{tab:rs_models}, early methods like ViT-based Scale-MAE~\cite{reed2023scale} and Cross Scale MAE~\cite{tang2023cross} directly borrowed natural image architectures, demonstrating strong transferability on large-scale datasets such as FMoW.

Additionally, SMLFR~\cite{dong2024generative} leveraged a ConvNeXt-Large backbone to capture long-range dependencies, while SpectralGPT~\cite{hong2024spectralgpt} introduced a Transformer-based architecture tailored for spectral spatial modeling. DyVis~\cite{chen2025dynamicvis} proposed a dynamic vision model that adapts to varying input resolutions and modalities, enhancing generalization across diverse remote sensing tasks.

\subsection{Multi-modal Pretraining}

With the increasing diversity of remote sensing data spanning spatial resolution, spectral bands, and imaging modalities, researchers have explored self-supervised pretraining for multi-modal data fusion. SpectralGPT~\cite{hong2024spectralgpt} employs a joint encoder-decoder framework to model RGB, depth, and semantic data. FG-MAE~\cite{wang2024feature} utilizes domain-specific features such as gradient histograms and NDVI as reconstruction targets, improving cross-modal consistency.

TOV~\cite{tao2023tov} adopts a two-stage pretraining strategy, transitioning from natural images to remote sensing data, achieving notable generalization gains. OFA-L~\cite{xiong2024one} establishes a unified remote sensing representation learning framework capable of joint multi-source data training. The DMC series models~\cite{wanyan2023dino}, combining ResNet or Swin Transformer backbones with DINO contrastive learning, achieve promising results on datasets like SeCo-100K.

Moreover, GeRSP~\cite{huang2024generic} introduces generic sensor-agnostic representations through a shared projection space, while PIS~\cite{an2024pretrain} focuses on invariant spectral modeling across heterogeneous sensors.

\subsection{Exponential Moving Average and Self Distillation Learning}
The Exponential Moving Average (EMA) mechanism plays a critical role in stabilizing self-supervised learning and improving feature consistency across training iterations~\cite{he2020momentum,assran2023self}. Originally introduced in MoCo v2~\cite{chen2020improved} for contrastive learning, EMA has since been widely adopted in frameworks like DINO~\cite{caron2021emerging} and I-JEPA~\cite{assran2023self}, where it is used to maintain a slowly updated teacher model that provides stable pseudo-labels or reconstruction targets for the student model.

In the context of multi-modal remote sensing image learning, EMA helps mitigate modality-specific noise and spatial-spectral inconsistencies, especially in ultra-high-resolution (UHR) settings. By maintaining a moving average of the student encoder's parameters, the teacher network generates more reliable cross-modal representations, which serve as stable supervision signals during training.

\subsection{Summary of Related Work}

In summary, the preceding sections have reviewed and analyzed recent advances in self-supervised learning for remote sensing imagery, particularly focusing on large-scale pretraining datasets, masked image modeling (MIM), multi-modal fusion strategies, and knowledge distillation mechanisms such as exponential moving average (EMA). These developments have significantly advanced the field by enabling more effective feature learning from unlabeled data across diverse modalities and spatial resolutions.

However, several critical challenges remain unaddressed. First, most existing pretraining datasets are limited to medium or low spatial resolution, which hinders their applicability to ultra-high-resolution (UHR) scene analysis. Second, pixel-level alignment between heterogeneous modalities (e.g., DOM and DSM) is often lacking, impeding precise cross-modal modeling. Third, current self-supervised frameworks fail to fully exploit modality-shared semantics while preserving modality-specific characteristics, especially under high-resolution settings where structural details are crucial for downstream tasks.

To address these limitations, we propose a novel framework termed \textit{Modality-Shared Self-supervised Distillation Framework} (\textbf{MSSDF}).  To support methodological development and benchmarking, we also introduce a new high-resolution multi-modal dataset named \textbf{HR-Paris}, consisting of over 640,000 centimeter-level pixel-aligned DOM–DSM image pairs.

\section{HR-Pairs Dataset}
\label{sec:hrdata}
\begin{figure}[htb]
    \centering
    \includegraphics[width=0.7\textwidth]{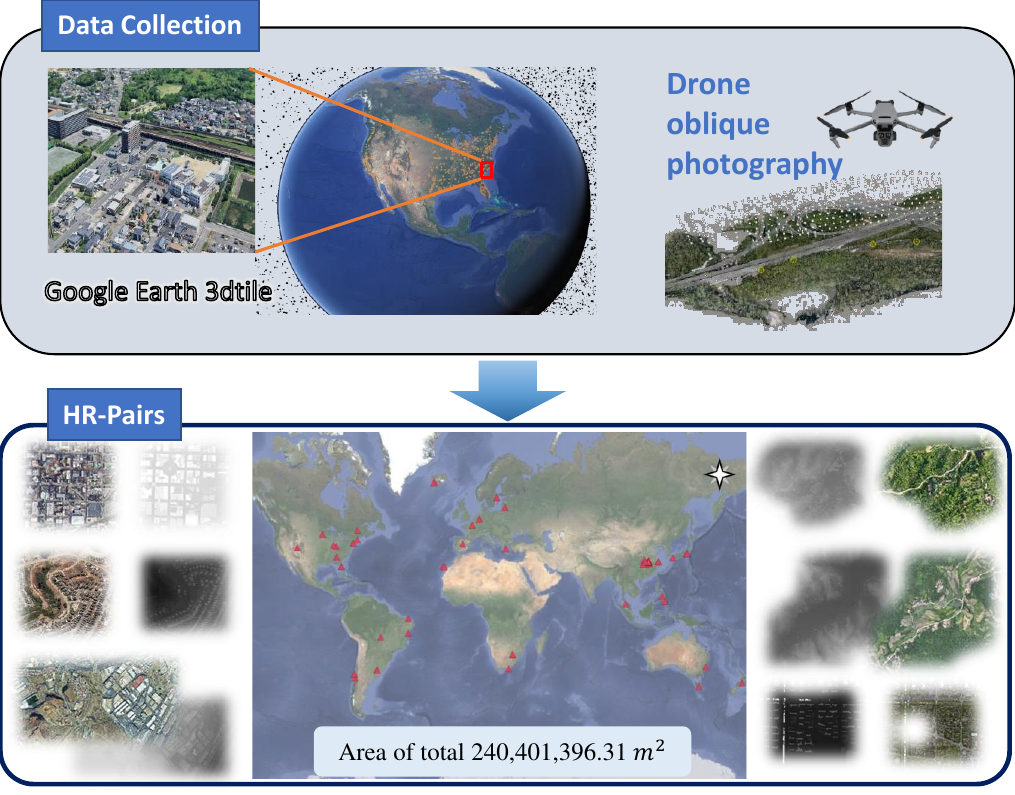}
    \caption{Construction pipeline of the HRPairs dataset. The red dots in the figure represent the data distribution areas. }
    \label{fig:dataset}
\end{figure}

To support high-resolution image pretraining tasks, we construct a dataset containing high-resolution DOM-DSM image pairs, namly HR-Paris. This dataset integrates Google Earth 3D Tiles and drone oblique photography data, covering a total area of 240,401,396.31 square meters. Aggregating multi-source data from over 80 medium and large cities worldwide, it encompasses diverse terrains ranging from plains to mountainous regions. Figure \ref{fig:dataset} presents an overview of the HRPairs dataset construction pipeline.

As shown in Figure~\ref{fig:dataset}, the raw data comprises two primary components: (1) globally available high-precision 3D models in tileset format , obtained from Google Earth; and (2) locally captured multi-view oblique imagery using drone platforms over selected target areas. This combination ensures complementary spatial coverage and local feature representation.

For the 3dtiles data, we perform orthographic projection to generate Digital Orthophoto Maps (DOMs) in 2D image space. Simultaneously, Digital Surface Models (DSMs) are extracted from depth information during the rendering process, ensuring geometric consistency with the corresponding DOMs. For the drone-acquired oblique imagery , we apply aerial triangulation and 3D reconstruction~\cite{zhang2023aerial,chen2024end} using ContextCapture software. The resulting DOM and DSM products are then rigorously calibrated to align with ground truth coordinates, achieving high geometric accuracy.

\begin{figure*}[htb]
    \centering
    \includegraphics[width=\textwidth]{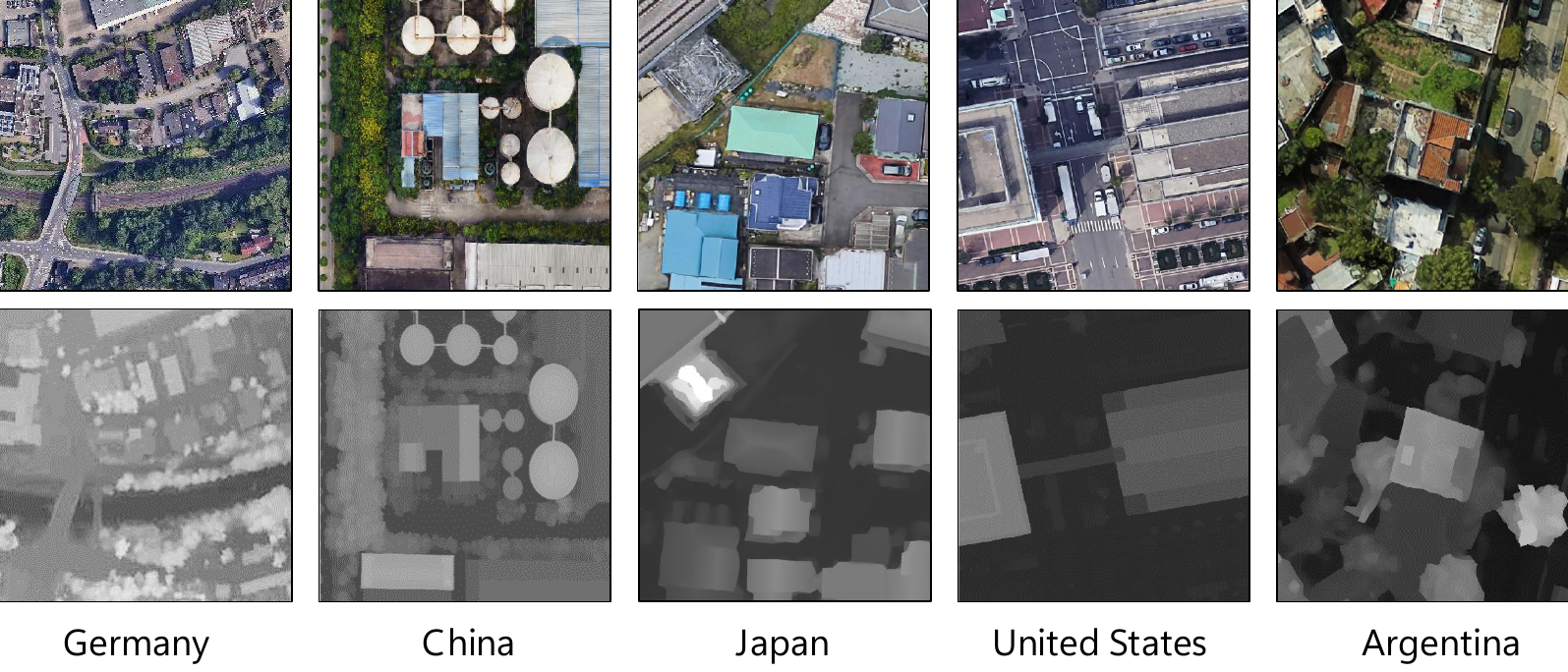}
    \caption{Visualization examples from HRPairs dataset.}
    \label{fig:dataset-demo}
\end{figure*}

The final dataset encompasses a wide range of terrain types and environmental conditions, including urban, rural, mountainous, and forested regions,  includes regions from over 10 countries and areas, including China, the United States, Canada, Japan, the United Kingdom, among others. To facilitate deep learning training and evaluation, we crop the original high-resolution images into non-overlapping patches of size 512 $\times$ 512 pixels , yielding approximately 640,000 paired DOM--DSM image patches. As illustrated in Figure~\ref{fig:dataset-demo}, each sample pair includes an RGB DOM image (first row) and its corresponding DSM representation (second row), visually highlighting the topographic variations across different landscapes.

\begin{figure*}[!ht]
    \centering
    \includegraphics[width=1\textwidth]{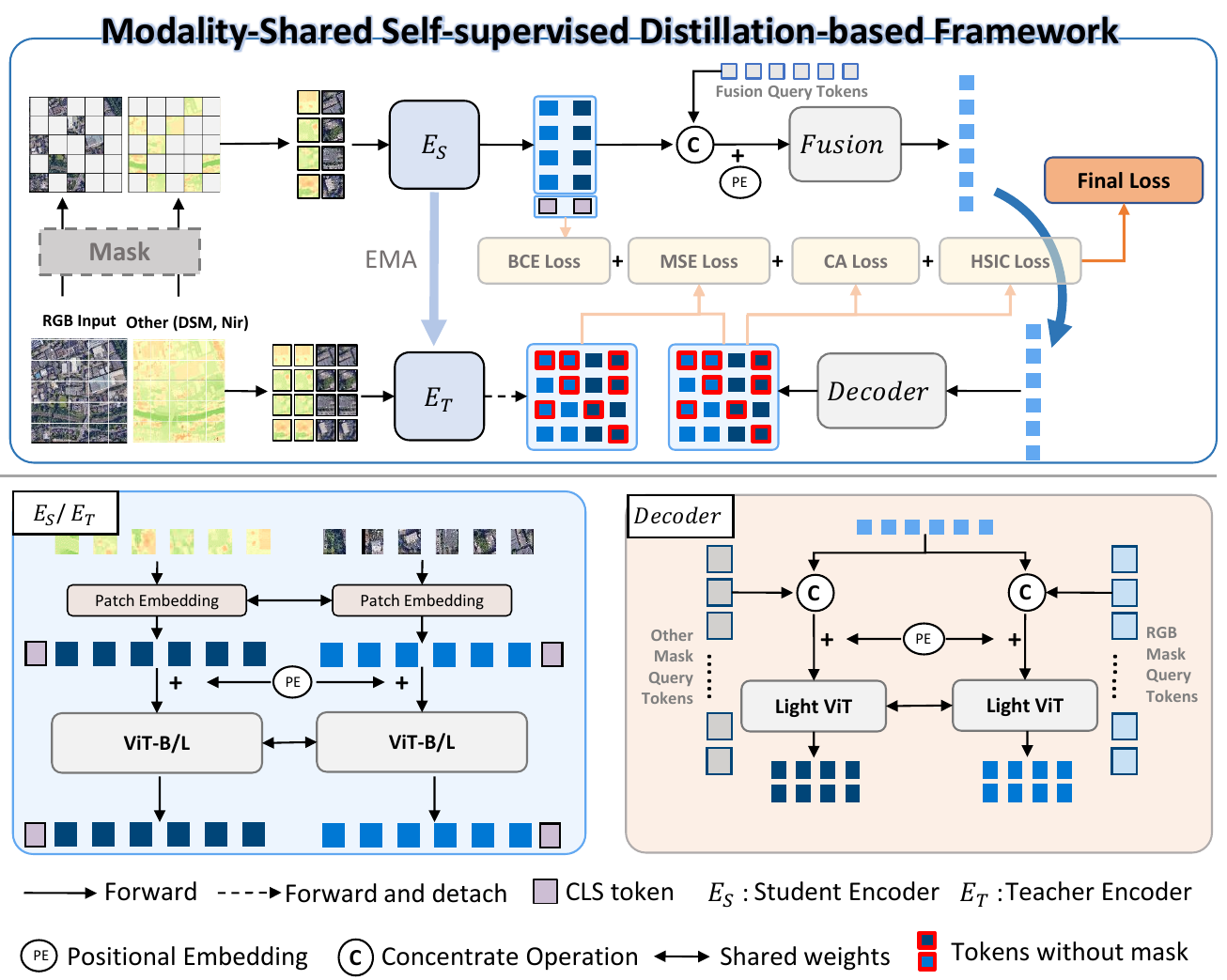}
    \caption{Modality-Shared Self-supervised Distillation-based Framework (MSSDF}
    \label{fig:model_arch}
\end{figure*}

\section{Methods}
\label{sec:method}
We propose a novel Modality-Shared Self-supervised Distillation Framework (\textbf{MSSDF}) designed to effectively exploit multi-modal remote sensing imagery. As illustrated in Figure~\ref{fig:model_arch}, MSSDF improves representation learning and generalization performance by integrating three key components: a modality-shared weight mechanism, self-supervised learning, and knowledge distillation. Central to the framework is a student-teacher architecture that incorporates both contrastive learning and knowledge distillation to facilitate effective feature extraction and knowledge transfer. This design seamlessly combines the advantages of multi-modal data processing with the strengths of self-supervised learning paradigms. In the following sections, we present a detailed description of the framework's architecture and operational principles.

\subsection{Model Architecture}
\label{sec:model_architecture}

As shown in Figure~\ref{fig:model_arch}, the proposed pre-training framework consists of four core components: the Student Encoder ($E_S$), the Teacher Encoder ($E_T$), the Fusion module, and the Decoder. The overall architecture is built upon Vision Transformers (ViT). The detailed configurations of each module are summarized in Table~\ref{tab:vit-b-config}.

The \textbf{Student Encoder} processes masked RGB inputs as well as other modality data. It is based on the standard ViT-B/16 or ViT-L/16 architecture, where images are divided into $16 \times 16$ patches and embedded into token vectors via a linear projection layer. Positional information is preserved using sinusoidal positional encoding. The \textbf{Teacher Encoder} shares the same architecture as the student encoder but receives unmasked inputs. Its parameters are updated via Exponential Moving Average (EMA) from the student encoder, which ensures stable target representations and enables the student encoder to gradually learn more robust feature representations.

\begin{table}[htbp]
\centering
\caption{Configuration of the Model Components.}
\label{tab:vit-b-config}
\begin{tabular}{l|ccc}
\hline
 & \textbf{ViT-B (Encoder)}  & \textbf{Fusion}& \textbf{Decoder}\\  
\hline 
Dim. & 768        & 384  & 384 \\ 
MLP Dim.     & 3072    & 1536 &  1536     \\ 
Layers   & 12    & 3  & 4        \\ 
Heads    & 12  &  6 & 6          \\ 
\# Param & $\sim$86M &  $\sim$4.2M & $\sim$5.6M         \\ 
\hline
\end{tabular}
\end{table}

\begin{figure}[h]
    \centering
    \includegraphics[width=0.5\textwidth]{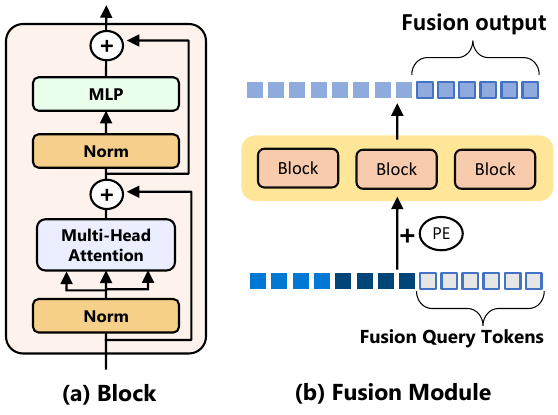}
    \caption{Fusion module architecture}
    \label{fig:fusion}
\end{figure}

The \textbf{Fusion module} integrates multi-modal features extracted by both encoders to generate a unified and discriminative feature representation. As shown in Figure~\ref{fig:fusion} and Table~\ref{tab:vit-b-config}, it is built upon a lightweight Transformer with 3 Transformer blocks.

\begin{figure}[htbp]
    \centering
    \includegraphics[width=0.4\textwidth]{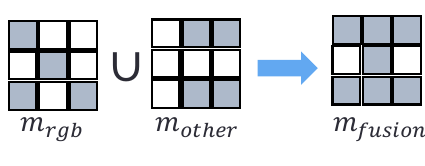}
    \caption{Position information of the output features from the Fusion module}
    \label{fig:fusionpe}
\end{figure}

The position information of the output features from the Fusion module is illustrated in Figure~\ref{fig:fusionpe}. In this figure, $ m_{\text{rgb}} $ and $ m_{\text{other}} $ represent the mask maps for RGB and another modality, respectively, while $ m_{\text{fusion}} $ denotes the mask map of the Fusion module, with dark regions indicating unmasked areas. If a patch remains unmasked in either the RGB or DSM modality, its fused feature will be retained in the output of the Fusion module. This ensures that even if certain regions are masked in some modalities, the model can still leverage information from the unmasked modalities for feature recovery and inference.

The \textbf{Decoder} is responsible for reconstructing the masked regions based on the fused features. As shown in Table \ref{tab:vit-b-config}, it also adopts a lightweight Transformer structure with 4 layers. During implementation, the decoder takes a set of fixed masked query tokens along with contextual features from the Fusion module to progressively recover the features of the masked parts. All four modules adopt standard Transformer layer logic, as shown in Figure~\ref{fig:fusion}(a).

\subsection{Modality-Aware Masking Strategy}

\subsubsection{Information-Aware Masking}
\label{sec:info_mask}
To preserve more informative content during the masking process, we propose an information-aware masking strategy that replaces traditional random masking with a content-dependent approach. Instead of uniformly sampling patches to mask, our method evaluates the information richness of each patch across different modalities and assigns adaptive masking probabilities accordingly.

\paragraph{\textbf{Patch-Level Information Measurement:}}
For each modality, we compute two simple yet effective metrics to measure the information density within each image patch:

- \textbf{Gradient Magnitude:} Reflects the presence of edges and texture details in the image. For a given patch $p$, its gradient magnitude is computed as Eq.~\ref{eq:grad}:
\begin{equation}\label{eq:grad}
    S_{\text{grad}}(p) = \frac{1}{N}\sum_{x \in p} \|\nabla I(x)\|_2
\end{equation}    
where $\nabla I(x)$ denotes the gradient vector at pixel $x$ computed using Sobel filters.

- \textbf{Local Variance:} Measures the intensity variation within a patch, indicating structural complexity, as shown in Eq.~\ref{eq:var}:
\begin{equation}\label{eq:var}
    S_{\text{var}}(p) = \frac{1}{N}\sum_{x \in p} (I(x) - \bar{I})^2
\end{equation}
where $\bar{I}$ is the mean intensity value over the patch.

These two metrics are combined linearly to form the final information score for each patch, as given in Eq.~\ref{eq:info}:
\begin{equation}\label{eq:info}
S(p) =S_{\text{grad}}(p) + S_{\text{var}}(p)
\end{equation}

\paragraph{\textbf{Mask Probability Calculation:}}
To handle multi-modal inputs (e.g., RGB+DSM or RGB+NirRG), we compute fused information score, $S_{\text{fusion}}$, for each patch by averaging the information scores from all modalities, as shown in Eq.~\ref{eq:fusion}:
\begin{equation}\label{eq:fusion}
    S_{\text{fusion}} = S_{\text{RGB}} +  S_{\text{other}}
\end{equation}

Based on the fused information score $S_{\text{fusion}}$, we adjust the masking probability $p_{\text{mask}}$ for each patch, as described in Eq.~\ref{eq:mask}:
\begin{equation}\label{eq:mask}
    p_{\text{mask}} =
    \begin{cases}
    0.8 & \text{if } S_{\text{fusion}} < Q_{20} \\
    0.3 & \text{if } S_{\text{fusion}} > Q_{80} \\
    0.5 & \text{otherwise}
    \end{cases}
\end{equation}
where $Q_{20}$ and $Q_{80}$ denote the 20th and 80th percentile values of $S_{\text{fusion}}$ within the current batch.

\subsubsection{Cross-Modal Masking Strategy}
\label{sec:cross_modal_masking}

To enhance the model's ability to learn cross-modal correlations during pre-training, we introduce a cross-modal masking strategy that builds upon the information-aware masking applied independently to each modality. This strategy encourages the model to explore relationships between modalities by selectively masking or substituting patches based on their preservation status in other modalities.

Given a pair of input modalities $(X_{\text{rgb}}, X_{\text{other}})$, we first apply modality-specific information-aware masking (as described in Section~\ref{sec:info_mask}) to generate masked versions $(\tilde{X}_{\text{rgb}}, \tilde{X}_{\text{other}})$. After this step, some patches are masked in both modalities according to their local information richness.

Subsequently, for each spatial position $p$, we perform the following Eq.~\ref{eq:cross_modal_masking}:

\begin{equation}
    \label{eq:cross_modal_masking}
\hat{X}_{\text{other}}(p) =
\begin{cases}
    \tilde{X}_{\text{other}}(p), & \text{w.p. } 1 - \rho, \\
    x \sim \mathcal{M}_{\text{other}}, & \text{w.p. } \rho,
\end{cases}
\end{equation}
where:

- $\rho \in [0,1]$ is the substitution probability. In our experiments, it starts at 0.1 and is gradually increased by 0.1 every 10 epochs, with a maximum value of 0.7. 
    
- $\mathcal{M}_{\text{other}}$ denotes the set of masked patches in $\tilde{X}_{\text{other}}$;
    
- $x \sim \mathcal{M}_{\text{other}}$ indicates random sampling from the masked region.

This mechanism introduces uncertainty into the input distribution and forces the model to reason across modalities when reconstructing missing information. It also avoids overfitting to co-occurring patches in aligned modalities, thereby improving robustness and generalization.

\subsection{Self-Supervised Training Objectives}
\label{sec:loss_function}

Our model is trained using a combination of self-supervised objectives designed to learn rich and discriminative multi-modal feature representations. The overall training objective consists of four complementary loss components: reconstruction loss, contrastive alignment loss, feature decorrelation loss, and auxiliary classification loss.

\subsubsection{Reconstruction Loss (\texttt{MSE Loss})}

We perform context-aware feature reconstruction in the latent space. Specifically, instead of using raw pixel values as targets, we adopt the token representations encoded by the Teacher Encoder at unmasked positions as the reconstruction targets. This design guides the Student Encoder to reconstruct masked regions based on high-level semantic features rather than low-level visual details.

Let $\mathcal{U}$ denote the set of unmasked positions and $\mathcal{M}$ the set of masked positions. Let $\mathbf{h}_t^{(p)}$ be the output token of the Teacher Encoder at position $p$, and let $\hat{\mathbf{h}}_s^{(p)}$ be the corresponding output of the Student Encoder through the Decoder. The reconstruction loss is defined as Eq.~\ref{eq:rec}:

\begin{equation}\label{eq:rec}
\mathcal{L}_{\text{rec}} = \frac{1}{|\mathcal{M}|} \sum_{p \in \mathcal{M}} \| \hat{\mathbf{h}}_s^{(p)} - \mathbf{h}_t^{(p)} \|_2^2
\end{equation}

This loss encourages the student model to recover feature representations consistent with those of the teacher model in the latent space, thereby learning more semantically meaningful and robust features.

This approach avoids potential noise introduced by pixel-level reconstruction and enables the model to focus on high-level structural consistency, aligning better with the goal of self-supervised representation learning.

\subsubsection{Contrastive Alignment Loss (\texttt{CA Loss})}

To enhance cross-modal consistency, we introduce a contrastive alignment loss that aligns semantically corresponding patches across different modalities. Given a query patch embedding $\mathbf{h}_i$, its positive pair $\mathbf{h}_i^+$ (e.g., the same spatial location in another modality), and a set of negative samples $\{\mathbf{h}_j^-\}$, the loss is defined as Eq.~\ref{eq:ca}:

\begin{equation}\label{eq:ca}
\mathcal{L}_{\text{align}} = -\frac{1}{N} \sum_i \log \frac{\exp(\mathbf{h}_i^\top \mathbf{h}_i^+ / \tau)}{\sum_j \exp(\mathbf{h}_i^\top \mathbf{h}_j^- / \tau)}
\end{equation}
where $\tau$ is a temperature scaling factor. This formulation pulls together aligned features while pushing apart mismatched ones, promoting robust cross-modal understanding.

\subsubsection{Feature Decorrelation via Linear HSIC (\texttt{HSIC Loss})}

To reduce redundancy among multi-modal features while maintaining computational efficient, we adopt a linear approximation of the Hilbert Schmidt Independence Criterion (HSIC)~\cite{marcou2016kernel}, referred to as \textbf{Linear HSIC}. This variant avoids the explicit computation of large kernel matrices and significantly reduces memory consumption, making it well-suited for large-scale self-supervised pre-training tasks.

Let $\mathbf{X} \in \mathbb{R}^{n \times d}$ and $\mathbf{Y} \in \mathbb{R}^{n \times d}$ denote the feature representations of two modalities at the same spatial positions, where $n$ is the number of patches and $d$ is the feature dimension. We first center the features by subtracting their respective means, as shown in Eq.~\ref{eq:center}:

\begin{equation} \label{eq:center}
    \mathbf{X}_c = \mathbf{X} - \mathbf{1}_n \left( \frac{1}{n} \mathbf{1}_n^\top \mathbf{X} \right), \quad
    \mathbf{Y}_c = \mathbf{Y} - \mathbf{1}_n \left( \frac{1}{n} \mathbf{1}_n^\top \mathbf{Y} \right)
\end{equation}
The Linear HSIC loss is then defined as Eq.~\ref{eq:hsic}:

\begin{equation} \label{eq:hsic}
\mathcal{L}_{\text{hsic}} = \frac{1}{(n - 1)^2} \| \mathbf{X}_c^\top \mathbf{Y}_c \|_F^2
\end{equation}
where $\| \cdot \|_F$ denotes the Frobenius norm.

This formulation encourages the model to learn statistically independent feature representations across modalities by minimizing the squared covariance between them.

\subsubsection{Auxiliary Classification Loss (\texttt{BCE Loss})}

We further introduce an auxiliary binary classification task to improve feature discriminability. In this task, the model is trained to predict whether a given patch belongs to a specific modality (e.g., RGB or DSM), without relying on human-annotated class labels.

Let $\mathbf{h}_i$ denote the feature vector of a patch, and $m_i \in \{0,1\}$ be the pseudo-label indicating its modality type. The auxiliary classification loss is defined as Eq.~\ref{eq:cls}:

\begin{equation}\label{eq:cls}
    \mathcal{L}_{\text{cls}} = -\frac{1}{N} \sum_i m_i \log \sigma(\mathbf{W}_c \mathbf{h}_i) + (1 - m_i) \log (1 - \sigma(\mathbf{W}_c \mathbf{h}_i))
\end{equation}
where $\sigma(\cdot)$ is the sigmoid function and $\mathbf{W}_c$ denotes the learnable classifier weights.

This auxiliary task encourages the model to learn more structured and separable feature spaces during pre-training.

\subsubsection{Overall Objective}

The final training objective combines all four loss terms, as shown in Eq.~\ref{eq:total}:

\begin{equation}\label{eq:total}
\mathcal{L}_{\text{total}} = \lambda_1 \mathcal{L}_{\text{rec}} + \lambda_2 \mathcal{L}_{\text{align}} + \lambda_3 \mathcal{L}_{\text{hsic}} + \lambda_4 \mathcal{L}_{\text{cls}}  
\end{equation}

where $\lambda_1, \lambda_2, \lambda_3, \lambda_4$ control the relative importance of each component. This composite loss function enables the model to learn both local and global feature representations that are accurate, consistent, complementary, and discriminative across modalities.

\section{Experiments and Results}
\label{sec:exp}
\subsection{Self-supervised Pre-training Setup}

The details of the settings are shown in Table~\ref{tab:pretrain}. 
The specific model architecture parameters are detailed in Table~\ref{tab:vit-b-config}. We initially perform pre-training on the \textit{fMoW} dataset and \textit{HRPairs} for 500 epochs.  On the \textit{fMoW} dataset, we adopt a multi-channel input comprising RGB and NIR-RG to better capture spectral features of land cover; whereas on the \textit{HRPairs} dataset, we utilize RGB combined with DSM (Digital Surface Model) as input modalities. Since DSM data typically contains a single band, it is replicated three times to satisfy the model's requirement for three-channel input.

\begin{table}
\small   
\centering
\caption{Ours Pre-training Settings}
\label{tab:pretrain}
\begin{tabular}{c|c|c}
\hline
\textbf{Name} & \textbf{Ours-B} & \textbf{Ours-L} \\
\hline
\textbf{Encoder} & \textbf{ViT-B} & \textbf{ViT-L} \\
\textbf{Input Size} & $320 \times 320$ & $320 \times 320$ \\
\textbf{Patch Size} & $16 \times 16$ & $16 \times 16$ \\
\textbf{Dropout} & $0.1$ & $0.1$ \\
\textbf{Drop Path} & $0.1$ & $0.1$ \\
\hline
\textbf{Init. Lr. Rate} & $1.5 \times 10^{-4}$ & $1.5 \times 10^{-4}$ \\
\textbf{Warmup Lr. Rate} & $1 \times 10^{-6}$ & $1 \times 10^{-6}$ \\
\textbf{Lr. Schedule} & Cosine Annealing & Cosine Annealing \\
\textbf{Batch Size} & $512$ & $512$ \\
\textbf{Warmup Epochs} & $30$ & $30$ \\
\textbf{Total Epochs} & $500$ & $500$ \\
\hline
\textbf{Loss Weights } &\makecell[c]{
  $\lambda_1 = 1$ \\
  $\lambda_2 = 0.5$ \\
  $\lambda_3 = 0.2$ \\
  $\lambda_4 = 0.1$
} & \makecell[c]{
  $\lambda_1 = 1$ \\
  $\lambda_2 = 0.5$ \\
  $\lambda_3 = 0.2$ \\
  $\lambda_4 = 0.1$
} \\

\hline

\end{tabular}

\end{table}

Regarding data augmentation strategies, we employ common image enhancement techniques such as random rotation, horizontal/vertical flipping, random resizing and cropping, aimed at improving the robustness and generalization capability of the model. The optimizer and other training parameters are set as follows: we use batch size of 512, implemented across 4 computing nodes, each equipped with 4 NVIDIA A100 GPUs, totaling 16 GPUs. Gradient accumulation is employed, with gradients accumulated over 2 iterations per update. All experimental tasks are submitted via the SLURM cluster management system. The complete pre-training process takes approximately 14 days.

\subsection{Downstream Tasks}

To comprehensively evaluate the generalization capability of our proposed multi-modal self-supervised pre-training framework in remote sensing tasks, we conduct extensive downstream task evaluations across several representative remote sensing datasets. These tasks encompass the primary application scenarios in remote sensing image analysis, including: Scene Classification (SC) \cite{chen2018training}, Semantic Segmentation (SS) \cite{chen2018symmetrical}, Multi-modal Semantic Segmentation (MS) \cite{zhang2025invitation}, Monocular Depth Estimation (DE)~\cite{lv2024attention}, Binary Change Detection (BCD) \cite{tan2024segment}, Semantic Change Detection (SCD)~\cite{yang2020semantic}, Horizontal Bounding Box Object Detection (HBB) \cite{zhang2019geospatial}, and Oriented Bounding Box Object Detection (OBB) \cite{zhu2022multi}. Table~\ref{tab:downstream} lists the downstream datasets used, along with their corresponding task types and input modalities.

\begin{table}[h]
\centering
\caption{Downstream datasets and tasks}
\label{tab:downstream}
\small
\begin{tabular}{ccc}
\hline
\textbf{Dataset} &  \textbf{Data Modality} & \textbf{Task}   \\
\hline
AID~\cite{xia2017aid} & RGB & SC \\
MAID~\cite{long2021creating} & RGB & SC \\
RSD46-WHU~\cite{xiao2017high} & RGB & SC \\
NWPU~\cite{cheng2017remote} & RGB & SC \\
Potsdam & MS\&DSM & SS,MS,DE \\
Vaihingen &  MS\&DSM & SS,MS,DE  \\
US3D~\cite{bosch2019semantic} & MS\&DSM & SS,MS,DE \\
WHU Building~\cite{ji2018fully} & RGB & SS \\
LEVIR-CD~\cite{chen2020spatial} & RGB & BCD  \\
HRCUS-CD~\cite{zhang2023aernet} & RGB & BCD \\
SECOND~\cite{yang2020semantic} & RGB & BCD,SCD  \\
DOTA-2.0~\cite{xia2021dota} & RGB & HBB,OBB  \\
DOTA-1.5~\cite{xia2018dota} & RGB  &HBB,OBB \\
DIOR~\cite{li2020object} & RGB & HBB,OBB  \\
SODA-A~\cite{cheng2023towards} & RGB & HBB,OBB \\
\hline
Total & - & 26\\
\hline
\end{tabular}

\end{table}

\subsubsection{Evaluation on Scene Classification Tasks}
\label{sec:scene_classification}

\begin{table*}[h]
\centering
\small
\caption{Top-1 Accuary on scene classification tasks. FT (20) denotes fine-tuning with 20\% of training samples, and LP (FULL) denotes linear probing using the full training set. The best results are marked in \textbf{bold}, and the second-best are \underline{underlined}.}
\label{tab:aidcls_results}
\begin{tabular}{l|cc|cc|cc|c}
\hline
\multirow{2}{*}{Method} & \multicolumn{2}{c|}{AID} & \multicolumn{2}{c|}{NWPU} & \multicolumn{2}{c|}{RSD46-WHU} & MAID \\
\cline{2-8}
& FT (20) & LP (FULL) & FT (20) & LP (FULL) & FT (20) & LP (FULL) & FT (FULL) \\ 
\hline
ViT-B-Random & 70.43  & --     & 76.60   & --      & 72.66   & --      & 69.34   \\
ViT-B-ImageNet & 76.95 & 37.68  & 81.11  & 42.20    & 78.22   & 36.50    & 76.87   \\
Scale-MAE   & 92.38  & \underline{82.98}  & 91.56  & 61.88   & 90.53   & 75.96   & \underline{88.27}   \\
CS-MAE      & \underline{93.85}  & 76.57  & 90.76  & 75.22   & \underline{92.93}   & \underline{80.37}   & 85.88   \\
PIS         & 92.20  & 77.42  & 92.04  & \underline{79.02}   & 92.25   & 68.94   & 87.92   \\
SMLFR       & 91.55  & 82.76  & 92.56  & 72.46   & 90.61   & 77.46   & 87.08   \\
SpectralGPT & 91.55  & 68.26  & 86.11  & 56.88   & 90.29   & 76.55   & 77.74   \\
OFA-L       & 87.95  & 71.80  & 92.82  & 75.81   & 87.11   & 61.61   & 84.38   \\
DyVis       & \underline{93.85}  & 74.76  & \textbf{93.96}  & 62.23   & 90.92   & 59.93   & 85.28   \\
\hline
Ours-B      & 92.15  & 78.14  & 91.64  & 73.58   & 90.53   & 77.71   & 87.08   \\
Ours-L      & \textbf{94.90}  & \textbf{83.13}  & \underline{93.84}  & \textbf{79.77}   & \textbf{93.47}   & \textbf{80.85}   & \textbf{89.17}   \\
\hline
\end{tabular}
\end{table*}

\paragraph{Dataset Description: }

To comprehensively evaluate the performance of various pre-trained models on remote sensing image understanding, we conduct experiments on four widely used scene classification datasets:

\textbf{AID}~\cite{xia2017aid} contains approximately 10,000 aerial images covering 30 land-use scene categories such as airports, parks, and bridges. The images are of size $600 \times 600$ pixels, with rich semantic information and significant intra-class variation, making it a commonly used benchmark for remote sensing classification tasks.

\textbf{NWPU-RESISC45}~\cite{cheng2017remote} consists of 45 scene categories, each containing 700 images of size $256 \times 256$ pixels, totaling over 31,000 images. It is a well-established dataset for evaluating the robustness and generalization ability of remote sensing image classification models.

\textbf{RSD46-WHU}~\cite{xiao2017high} includes 46 typical scene classes such as urban areas, farmland, and lakes, with high-resolution and semantically clear images, suitable for fine-grained scene recognition tasks.

\textbf{Million-AID}~\cite{long2021creating} is a large-scale remote sensing dataset containing over one million images. Due to the lack of publicly available ground truth labels for its test set, we re-split the original training set into training, validation, and test subsets for model evaluation.

\begin{figure*}[ht]
    \centering
    \subcaptionbox{RSD46-WHU\label{fig:cfm_rsd46}}{
        \includegraphics[width=0.31\textwidth]{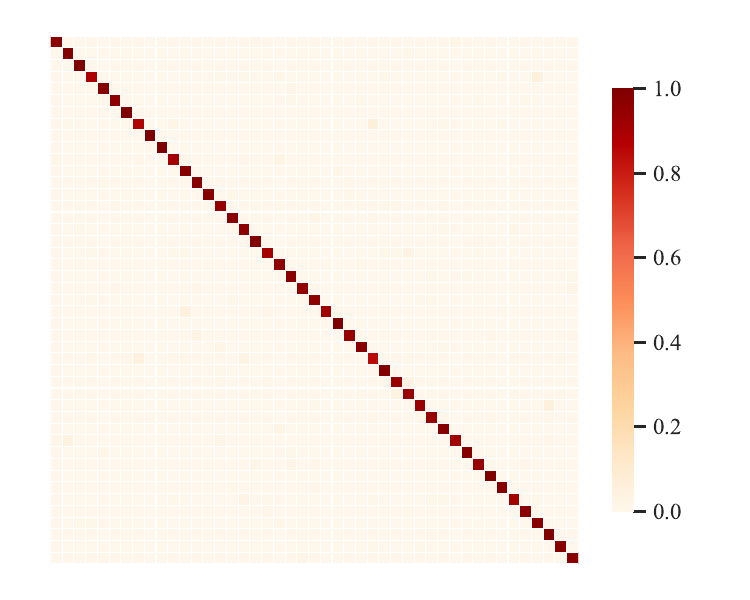}
    }
    \hfill
    \subcaptionbox{AID\label{fig:cfm_aid}}{
        \includegraphics[width=0.31\textwidth]{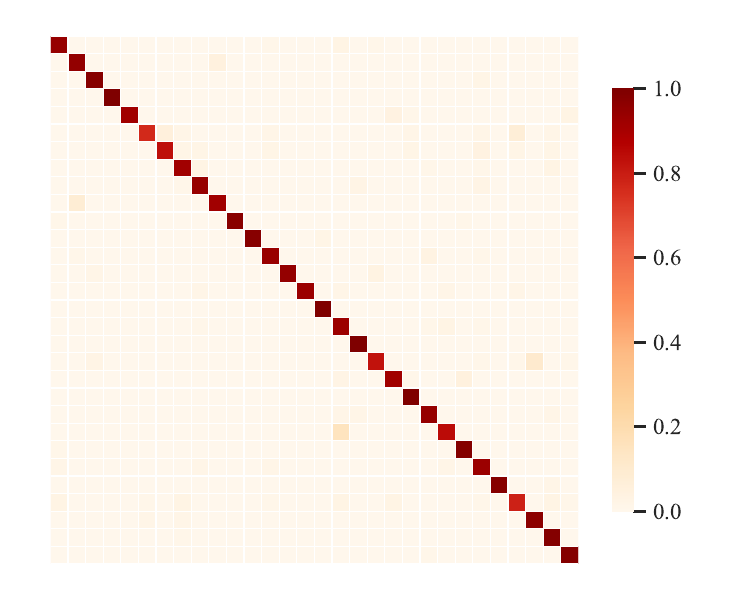}
    }
    \hfill
    \subcaptionbox{NWPU\label{fig:cfm_NWPU}}{
        \includegraphics[width=0.31\textwidth]{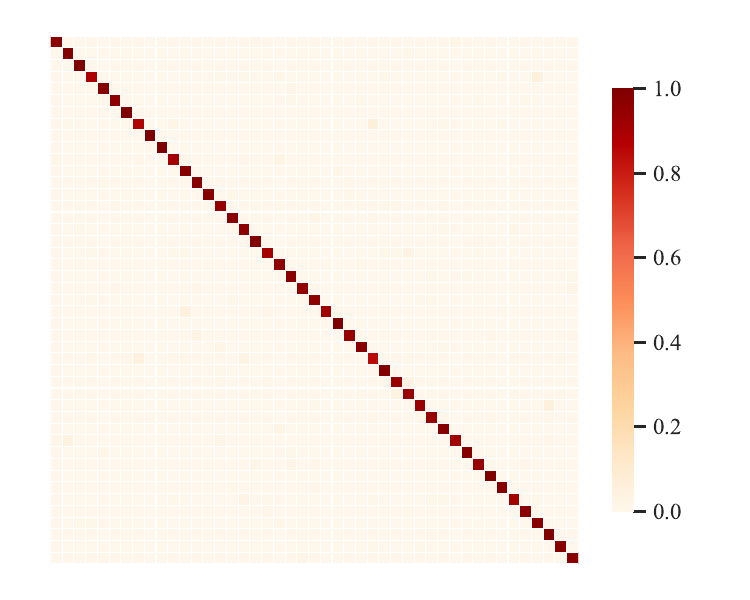}
    }
    \caption{Confusion matrix of our proposed method (Ours-L) on different datasets. High-definition images, along with their corresponding category labels, are provided in the supplementary materials Figure 1 to Figure 3.}
    \label{fig:confusion_matrices}
\end{figure*}

\paragraph{Experimental Settings:}
To comprehensively evaluate the effectiveness of the pre-trained models, we follow standard transfer learning protocols and consider both fine-tuning and linear probing settings. \textbf{FT} denotes \textit{fine-tuning}, where the entire model is trained using either 20\% (\textbf{FT (20)}) or all (\textbf{FT (FULL)}) of the labeled data. \textbf{LP} stands for \textit{linear probing}, where only a linear classifier is trained on top of frozen features, using the full training set (\textbf{LP (FULL)}). Performance is measured in terms of \textbf{Top-1 accuracy}.

\paragraph{Results and Analysis:}
The results are summarized in Table~\ref{tab:aidcls_results}. As shown in the table, our proposed methods (Ours-B and Ours-L) outperform existing state-of-the-art approaches on most metrics. Specifically, Ours-L achieves the best performance across all three main datasets under the FT (20) setting, with accuracy scores of \textbf{94.90\%}, \textbf{93.84\%}, and \textbf{93.47\%} on AID, NWPU, and RSD46-WHU, respectively. It also achieves the highest linear probing accuracies of \textbf{83.13\%}, \textbf{79.77\%}, and \textbf{80.85\%}, demonstrating strong feature generalization capabilities. The overall average accuracy across datasets (MAID) reaches \textbf{89.17\%}, surpassing other methods.

To further analyze the classification performance of our method, we visualize the confusion matrices for Ours-L on RSD46-WHU, AID, and NWPU in Figures~\ref{fig:cfm_rsd46}, \ref{fig:cfm_aid},  and \ref{fig:cfm_NWPU}, respectively. These matrices illustrate the model's prediction distribution across different scene categories, revealing high diagonal values that indicate strong classification accuracy. Some off-diagonal patterns highlight potential confusion between visually similar scenes (e.g., commercial area vs. medium residential ), suggesting room for improvement through more discriminative feature learning or attention mechanisms tailored for fine-grained distinctions.

In summary, our method demonstrates superior performance under limited label conditions and exhibits strong linear separability and cross-dataset generalization ability.

\subsubsection{Evaluation on Semantic Segmentation}

\paragraph{Dataset Description:}
We conduct experiments on four widely used remote sensing image datasets for semantic segmentation:

The \textbf{Potsdam} and \textbf{Vaihingen} datasets, published by ISPRS, consist of multispectral aerial images (infrared, red, green) and digital surface model, from which we use the NirRG bands (infrared, red, green) and DSM for evaluation. Both datasets are annotated with six semantic classes including impervious surfaces, buildings, low vegetation, trees, cars, and clutter.

The \textbf{US3D}~\cite{bosch2019semantic} dataset provides large-scale urban coverage with RGB, NIR, and DSM data. Following common practice, we use the NirRG bands and DSM for evaluation. The \textbf{WHU Building}~\cite{ji2018fully} dataset released by Wuhan University focuses specifically on building extraction and includes RGB images and binary annotations indicating building presence.

To evaluate the generalization ability under limited supervision, \textit{we randomly select only 50\% of the original training samples for fine-tuning and validation across all datasets}. This setting reflects realistic scenarios where labeled data is scarce and emphasizes the importance of strong pre-training for downstream tasks.

\paragraph{Semantic Segmentation Framework and Multi-Modal Fusion Strategy:}
In this work, we adopt \textbf{UPerNet}~\cite{xiao2018unified} as the semantic segmentation framework. UPerNet is an encoder-decoder architecture widely used in remote sensing image analysis due to its effective integration of multi-scale contextual information via the \textit{Pyramid Pooling Module (PPM)}.

To handle multi-modal inputs (e.g., NirRG and DSM), we employ a straightforward yet effective fusion strategy. Specifically, both modalities are first independently processed through the same pre-trained vision transformer backbone to extract their respective feature maps. These features are then concatenated channel-wise at multiple network depths, allowing the model to learn joint representations that exploit complementary information from both modalities. The fused features are subsequently fed into the UPerNet decoder for pixel-wise semantic prediction. Performance is measured in terms of $mIoU$.

\begin{table*}[htb]
\centering
\small
\caption{Semantic Segmentation $mIoU$ results on Potsdam, Vaihingen, US3D, and WHU Building datasets. The best results are marked in \textbf{bold}, and the second-best are \underline{underlined}.}
\label{tab:potsdam_seg_results_dummy}
\begin{tabular}{l|cc|cc|cc|c}
\hline
\multirow{2}{*}{\textbf{Method}} & \multicolumn{2}{c|}{\textbf{Potsdam}} &  \multicolumn{2}{c|}{\textbf{Vaihingen}} & \multicolumn{2}{c|}{\textbf{US3D}} & \textbf{WHUB} \\
\cline{2-8}
& NirRG & NirRG-DSM & NirRG & NirRG-DSM & NirRG & NirRG-DSM & RGB \\
\hline
ViT-B-Random & 67.10 & 67.40 & 64.44 & 65.06 & 57.92 & 63.75 & 84.78 \\
ViT-B-ImageNet & 69.65 & 70.63 & 65.34 & 66.58 & 60.36 & 66.24 & 85.66 \\
Scale-MAE & 78.35 & 80.91 & \underline{71.44} & \underline{73.06} & 67.03 & 73.79 & 90.62 \\
CS-MAE & 78.35 & \underline{81.57} & 71.37 & 71.40 & 66.44 & \underline{74.55} & 89.58 \\
PIS & 70.77 & 76.19 & 68.85 & 70.18 & 61.85 & 69.83 & 88.42 \\
SMLFR & 76.22 & 73.64 & 69.58 & 71.68 & 65.21 & 72.10 & 89.91 \\
SpectralGPT & 75.75 & 78.49 & 70.01 & 72.37 & 63.73 & 71.67 & 86.95 \\
OFA-L & \underline{78.36} & 81.42 & 70.81 & 72.38 & \underline{67.41} & 73.05 & \underline{91.59} \\
DyVis & 76.63 & 79.22 & 67.84 & 67.63 & 65.42 & 72.63 & 90.70 \\
\hline
Ours-B & 77.42 & 80.18 & 70.11 & 72.70 & 66.08 & 72.44 & 90.94 \\
Ours-L & \textbf{79.81} & \textbf{83.03} & \textbf{72.04} & \textbf{74.54} & \textbf{68.12} & \textbf{75.24} & \textbf{91.79} \\
\hline
\end{tabular}
\end{table*}

\paragraph{Results and Analysis:}

\begin{figure*}[htb]
\centering
\includegraphics[width=\linewidth]{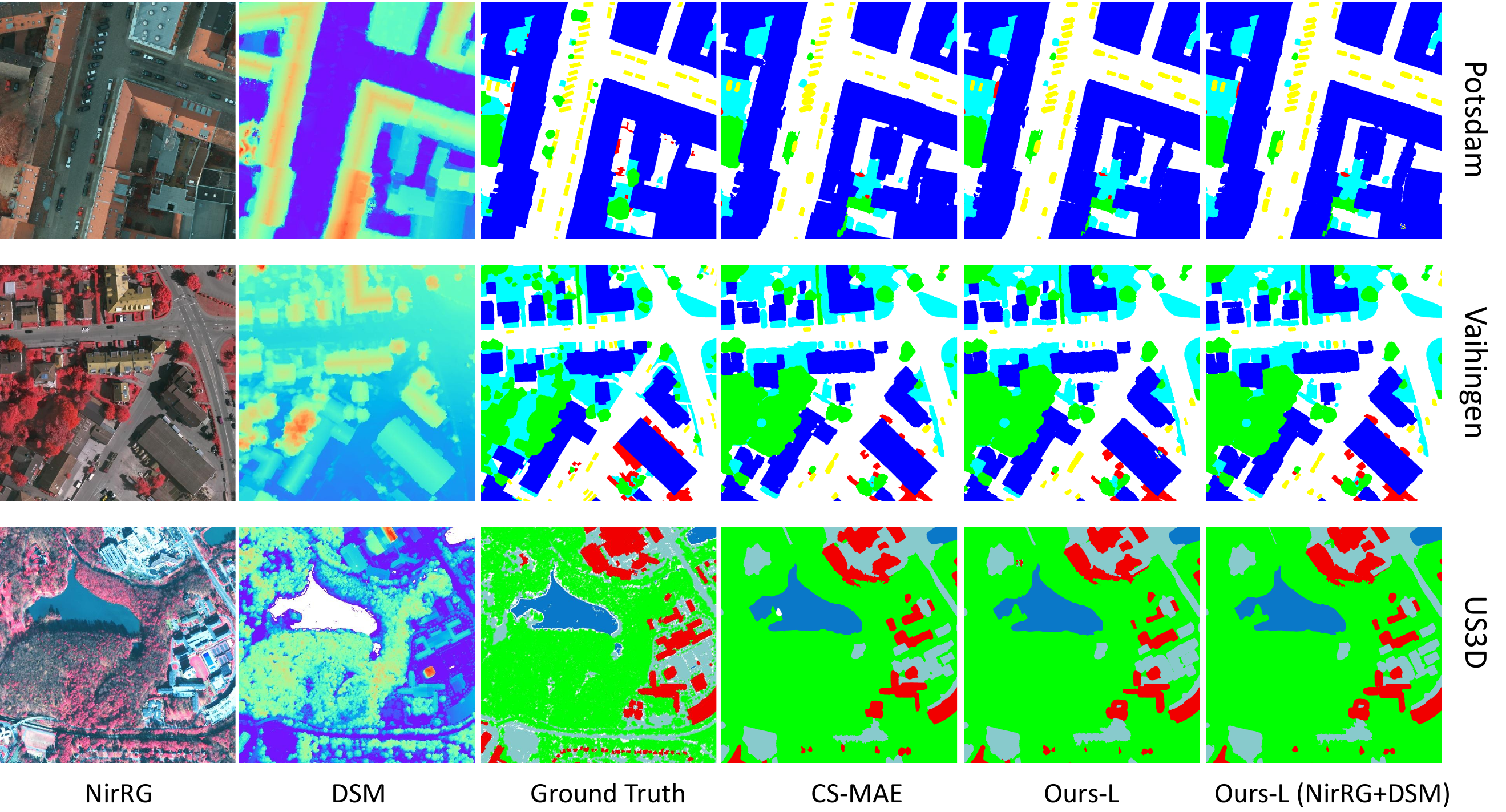}
\caption{Qualitative results on Potsdam, Vaihingen, and US3D datasets.}
\label{fig:seg_results_pot}
\end{figure*}

As shown in Table~\ref{tab:potsdam_seg_results_dummy}, our methods achieve consistently superior performance across different input modalities. 

On the \textbf{Potsdam} dataset: With the NirRG modality, Ours-B achieves an mIoU of \textbf{77.42}, while Ours-L reaches \textbf{79.81}, outperforming existing self-supervised and supervised pre-training approaches. When DSM is incorporated, Ours-L further improves to \textbf{83.03 mIoU}. On the \textbf{Vaihingen} dataset: Ours-L achieves \textbf{72.04 mIoU} with NirRG and \textbf{74.54 mIoU} with NirRG-DSM, surpassing previous state-of-the-art methods such as Scale-MAE and OFA-L. indicating its strong capability in multi-modal feature fusion.

For the \textbf{US3D} dataset: Ours-L obtains \textbf{68.12 mIoU} using NirRG and \textbf{75.24 mIoU} when fused with DSM, both setting new state-of-the-art results. These improvements are particularly evident in areas with significant terrain variations and complex height distributions.

\begin{figure*}[htb]
\centering
\includegraphics[width=\textwidth]{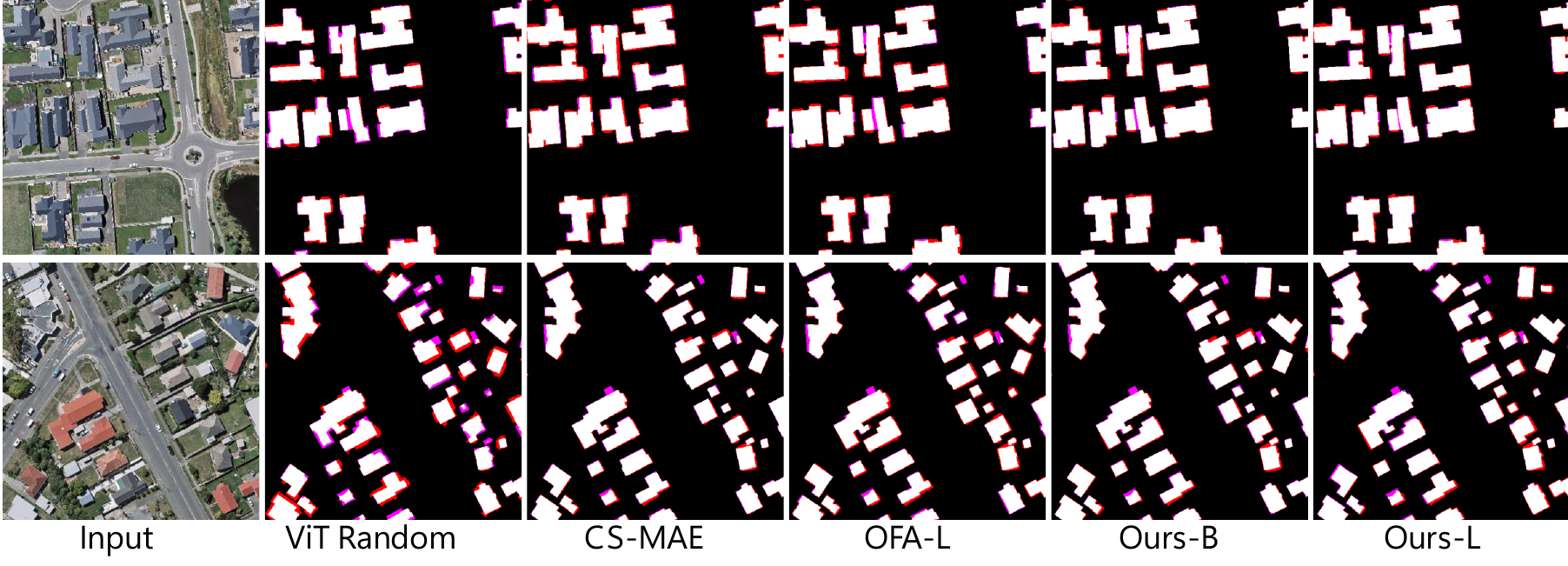}
\caption{Qualitative results on WHU Building datasets. \textcolor{red}{Red} denotes \textbf{False Positives} (background predicted as building), and \textcolor[rgb]{1,0,1}{magenta} represents \textbf{False Negatives} (building predicted as background).}
\label{fig:seg_results_whub}
\end{figure*}

On the \textbf{WHU Building} dataset using RGB images: Ours-L achieves an impressive \textbf{91.79 mIoU}, slightly surpassing OFA-L (91.59) and significantly outperforming ImageNet-pretrained ViT-B. This highlights the strong transferability of our pre-trained representations to RGB-based building extraction tasks.

In addition to quantitative results, we also provide qualitative comparisons as shown in Figure~\ref{fig:seg_results_pot} and~\ref{fig:seg_results_whub}. Visual inspection reveals that our method produces more accurate and spatially coherent segmentation maps, especially in densely built-up areas, shadowed regions, and low-contrast boundaries. Compared to baselines such as CS-MAE, our approach exhibits fewer false positives and negatives, demonstrating enhanced robustness and contextual understanding.

In summary, the proposed pre-training strategy combined with the UPerNet segmentation framework enables strong performance across diverse remote sensing datasets and modalities, even under a limited-data regime.

\subsubsection{Evaluation on Change Detection Tasks}

\paragraph{Datasets Description:}

We evaluate our pre-training on three change detection datasets:

- \textbf{LEVIR-CD}~\cite{chen2020spatial} is a large-scale high-resolution dataset focusing on building changes. It contains 637 pairs of $1024 \times 1024$ pixel Google Earth images with a spatial resolution of 0.5 m/pixel, spanning 5-14 years.

- \textbf{HRCUS-CD}~\cite{zhang2023aernet} targets complex urban scenes with building changes, including over 11,000 image pairs of $256 \times 256$ pixels at 0.5 m/pixel resolution. It covers challenging conditions such as vegetation interference and industrial zones.

- \textbf{SECOND}~\cite{yang2020semantic} is a multi-semantic change detection dataset containing 4,662 aerial image pairs from multiple cities. Each image is $512 \times 512$ pixels, annotated at the pixel level for six common land-cover categories (e.g., buildings, trees, water bodies).

To evaluate the generalization ability under limited supervision, \textit{we randomly select only 50\% of the original training samples for fine-tuning and validation across all datasets}. The performance is evaluated using the mIoU metric.

\paragraph{Change Detection Framework:}

We adopt the \texttt{SiamEncoderDecoder} framework for change detection, which is a widely used paradigm for modeling bi-temporal remote sensing images. In this architecture, two branches share weights to extract features from the ``pre-change'' and ``post-change'' image pairs independently. The extracted feature maps are then concatenated or differenced at multiple levels to capture discriminative change patterns. For the decoder component, we employ the \textbf{ChangeStarHead}~\cite{zheng2021change}, a lightweight yet effective head specifically designed for change detection tasks.

\paragraph{Results and Analysis:}

\begin{table*}[htb]
\centering
\caption{Change Detection $mIoU$ Results on LEVIR-CD, HRCUS-CD, and SECOND. The best results are marked in \textbf{bold}, and the second-best are \underline{underlined}.}
\label{tab:cd_results}
\begin{tabular}{lcccc}
\hline
\textbf{Method} & \textbf{LEVIRE-CD} & \textbf{HRCUS-CD} & \textbf{SECOND-BCD} & \textbf{SECOND-SCD}\\
\hline
ViT-B-Random & 71.41 & 55.15 & 65.24 & 34.45 \\
ViT-B-ImageNet & 73.56 & 56.80 & 67.51 & 39.18 \\
Scale-MAE & 84.58 & 61.85 & 71.59 & 47.44 \\
CS-MAE & 83.06 & \textbf{62.62} & \underline{71.83} & \underline{44.47} \\
PIS & 78.71 & 59.15 & 68.42 & 40.96 \\
SMLFR & 80.15 & 59.20 & 68.73 & 41.66 \\
SpectralGPT & 81.28 & 60.91 & 68.88 & 41.10 \\
OFA-L & \underline{84.79} & 61.18 & 71.37 & 45.51 \\
DyVis & 82.81 & 61.68 & 70.83 & 45.55 \\
\hline
Ours-B & 83.09 & 61.86 & 71.81 & 47.07 \\
Ours-L & \textbf{85.90} & \underline{62.21} & \textbf{72.01} & \textbf{47.51} \\
\hline
\end{tabular}

\end{table*}

Table~\ref{tab:cd_results} presents a comprehensive comparison between our method and state-of-the-art self-supervised and supervised approaches across all three datasets.  On \textbf{LEVIR-CD}, our method achieves an F1-score of \textbf{83.09\%} for Ours-B and \textbf{85.90\%} for Ours-L, outperforming recent methods such as OFA-L (84.79\%), and  DyVis (82.81\%).  On the more challenging \textbf{HRCUS-CD} dataset, our model also shows strong performance, achieving \textbf{61.86\%} and \textbf{62.21\%} mIoU for Ours-B and Ours-L respectively.    
    
On \textbf{SECOND}, our method excels in both binary and semantic change detection, achieving \textbf{72.01\%} on SECOND-BCD and \textbf{47.51\%} on SECOND-SCD with Ours-L. These results outperform methods such as SpectralGPT, OFA-L, indicating robustness in handling diverse land-cover change patterns.

\begin{figure*}[!hbt]
    \centering
    \includegraphics[width=1\textwidth]{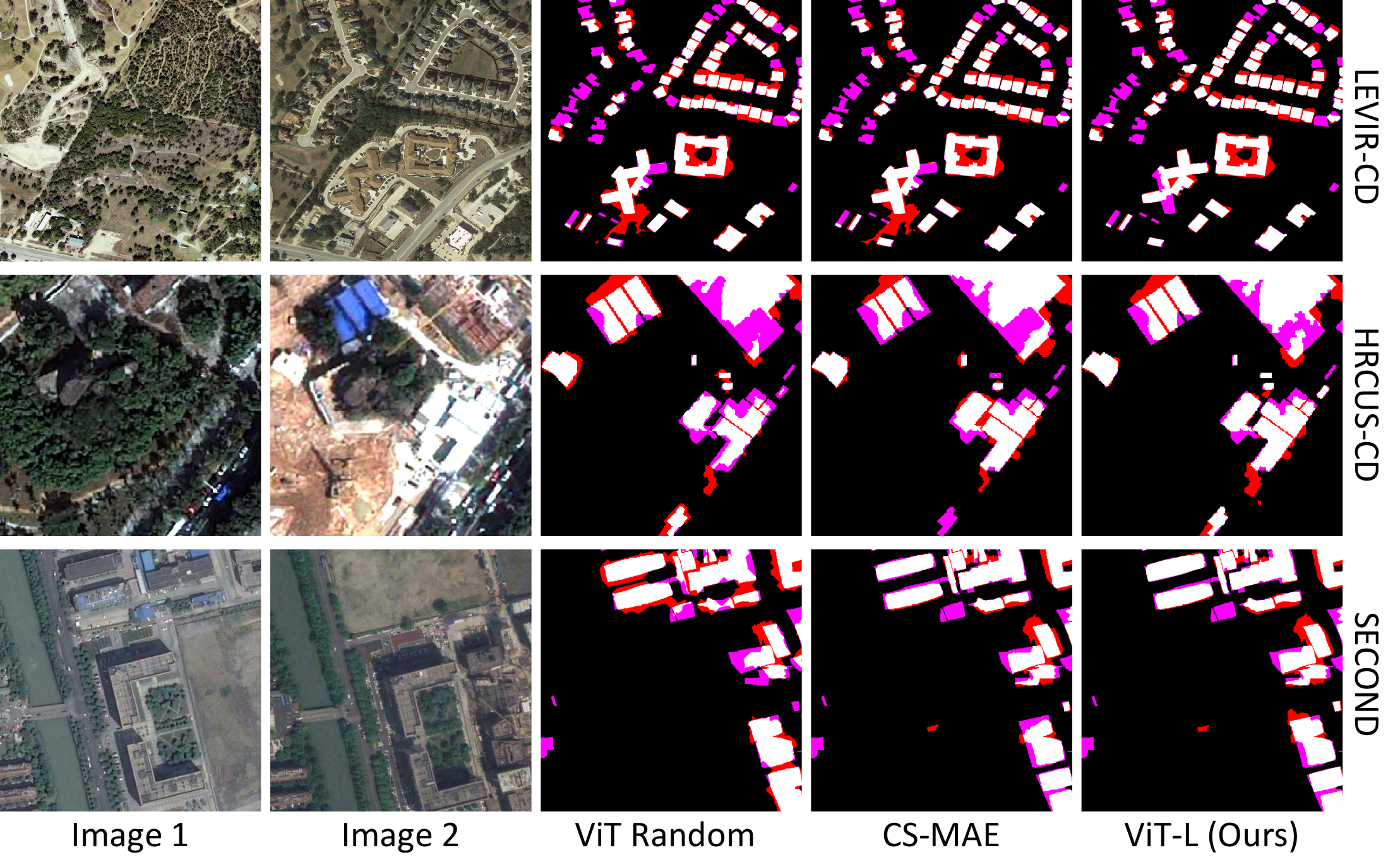}
    \caption{Binary Change Detection Results on LEVIR-CD, HRCUS-CD, and SECOND. \textcolor{red}{Red} denotes \textbf{False Positives} (non-change predicted as change), and \textcolor[rgb]{1,0,1}{magenta} represents \textbf{False Negatives} (change predicted as non-change).}
    \label{fig:cd_results}
\end{figure*}

In addition to quantitative metrics, we also provide visual comparisons of change maps generated by different methods on selected samples from LEVIR-CD and HRCUS-CD. As shown in Figure~\ref{fig:cd_results}, our method produces more accurate and complete change regions, especially in complex scenarios involving partial occlusions, similar-looking structures, or small-scale changes. Compared to other self-supervised and ImageNet-pretrained baselines, our model better preserves object boundaries and reduces false alarms caused by spectral inconsistencies or environmental noise.

\subsubsection{Evaluation on Object Detection Tasks}

\paragraph{Dataset Description:}
We evaluate our method on the rotation-aware object detection task using four widely used remote sensing benchmarks: \textbf{DOTA-v1.5}~\cite{xia2018dota}, \textbf{DOTA-v2.0}~\cite{xia2021dota}, \textbf{SODA-A}~\cite{cheng2023towards}, and \textbf{DIOR}~\cite{li2020object}. \textbf{DOTA} contains diverse object categories such as planes, ships, vehicles, and bridges in complex urban and rural scenes. \textbf{SODA-A} is a large-scale benchmark specialized for small object detection task under aerial scenes, which has 800203 instances with oriented rectangle box annotation across 9 classes. \textbf{DIOR} focuses on object detection at multiple scales with a wide coverage of object classes.

These datasets contain high-resolution aerial images annotated with both horizontal bounding boxes (HBB) and oriented bounding boxes (OBB), enabling evaluation of detection performance under arbitrary object orientations. Only \textit{50\% of the training samples are used} across all datasets during fine-tuning, emphasizing the importance of pre-training.

\paragraph{Model Architecture:}
For object detection, we adopt two widely used frameworks:

- \textbf{Horizontal bounding box (HBB)} detection: We use the \textbf{Faster R-CNN}~\cite{ren2016faster} architecture with a vision transformer backbone (Ours-B or Ours-L).
    
- \textbf{Oriented bounding box (OBB)} detection: We employ \textbf{Oriented R-CNN}~\cite{xie2021oriented}, which extends Faster R-CNN to handle rotated objects by introducing angle prediction and rotated RoI operations.

The feature pyramid network (FPN) is used to extract multi-scale features, and the pre-trained vision transformer weights are fine-tuned end-to-end on each downstream detection dataset. We report results using the standard mean average precision metric: \textbf{mAP} for HBB detection and OBB detection

\begin{table*}[htb]
\centering
\caption{Object Detection $mAP$ Results on DOTA, SODA-A, and DIOR. Both the HBB and OBB tasks are include. The best results are marked in \textbf{bold}, and the second-best are \underline{underlined}.}
\label{tab:object_results}
\begin{tabular}{l|cc|cc|cc|cc}
\hline
\multirow{2}{*}{\textbf{Method} }& \multicolumn{2}{c|}{\textbf{DOTAv1.5}} & \multicolumn{2}{c|}{\textbf{DOTAv2}} & \multicolumn{2}{c|}{\textbf{SODA-A}} & \multicolumn{2}{c}{\textbf{DIOR}} \\
\cline{2-9}
& \textbf{HBB} & \textbf{OBB} & \textbf{HBB} & \textbf{OBB} & \textbf{HBB} & \textbf{OBB} & \textbf{HBB} & \textbf{OBB} \\
\hline
ViT-B-Random & 61.24 & 54.35 & 46.71 & 30.23 & 50.11 & 36.33 & 48.30 & 24.71 \\
ViT-B-ImageNet & 62.01 & 54.59 & 48.46 & 31.60 & 50.87 & 36.95 & 50.13 & 25.45 \\
Scale-MAE & 66.65 & 61.03 & 50.64 & 37.11 & 52.12 & 43.20 & 53.82 & 29.68 \\
CS-MAE & 67.93 & \textbf{62.78} & 52.90 & 38.19 & \underline{53.37} & \underline{45.64} & 55.20 & 31.25 \\
SpectralGPT & 66.65 & 60.03 & 50.64 & 37.11 & 52.12 & 43.20 & 53.82 & 29.68 \\
OFA-L & \underline{68.29} & 61.62 & \textbf{54.10} & \underline{38.57} & 53.17 & 45.22 & \textbf{55.81} & \underline{32.76} \\
DyVis & 67.54 & 61.94 & 51.61 & 37.78 & 53.34 & 44.18 & 55.00 & 30.89 \\
\hline
Ours-B & 67.93 & 61.78 & 51.90 & 38.19 & \underline{53.37} & 44.64 & 55.28 & 31.25 \\
Ours-L & \textbf{68.83} & \underline{62.00} & \underline{53.98} & \textbf{39.59} & \textbf{53.85} & \textbf{45.90} & \underline{55.61} & \textbf{33.62} \\
\hline
\end{tabular}
\end{table*}

\paragraph{Results and Analysis:}
Table~\ref{tab:object_results} summarizes the performance of various pre-training strategies on both HBB and OBB object detection tasks across four remote sensing datasets: DOTAv1.5, DOTAv2, SODA-A, and DIOR. Our proposed methods, \textbf{Ours-B} and \textbf{Ours-L}, consistently outperform existing self-supervised and supervised approaches.

On the \textbf{DOTAv1.5} dataset, \textbf{Ours-L} achieves a HBB mAP of \textbf{68.83} and an OBB mAP of \textbf{62.00}, surpassing all compared methods including CS-MAE and OFA-L.  For the more challenging \textbf{DOTAv2} dataset, \textbf{Ours-L} achieves \textbf{53.98} mAP and \textbf{39.59} mAP in HBB and OBB task, respectively. On the \textbf{SODA-A} dataset, Ours-L achieves \textbf{53.85} HBB mAP and \textbf{45.90} OBB mAP, setting a new state-of-the-art result. On the \textbf{DIOR} dataset,  Ours-L reaches \textbf{55.62} mAP and \textbf{33.62} mAP, significantly outperforming ViT-B/ImageNet, highlighting its strong transferability across different remote sensing domains.

\begin{table*}[htb]
\centering

\caption{Depth estimation results on Potsdam, Vaihingen and US3D. The best results are marked in \textbf{bold}, and the second-best are \underline{underlined}.}
\label{tab:depth_results}
\small
\resizebox{\textwidth}{!}{
\begin{tabular}{l|ccc|ccc|ccc}
\hline
\multirow{2}{*}{\textbf{Method}} & \multicolumn{3}{c|}{\textbf{Potsdam}} &  \multicolumn{3}{c|}{\textbf{Vaihingen}} & \multicolumn{3}{c}{\textbf{US3D}} \\
\cline{2-10} 
 & \textbf{D1 $\uparrow$} & \textbf{RMSE $\downarrow$ }  & \textbf{SiLog $\downarrow$}  & \textbf{D1 $\uparrow$} & \textbf{RMSE $\downarrow$ }  & \textbf{SiLog $\downarrow$}  & \textbf{D1 $\uparrow$} & \textbf{RMSE $\downarrow$ }  & \textbf{SiLog $\downarrow$}  \\
\hline
ViT-B-Random & 38.10 & 0.167 & 0.789 & 33.43 & 0.183 & 0.564 & 14.58 & 0.250 & 1.687 \\
ViT-B-ImageNet & 39.88 & 0.160 & 0.702 & 33.59 & 0.185 & 0.573 & 14.71 & 0.244 & 1.619 \\
Scale-MAE & 59.79 & 0.090 & 0.421 & 42.50 & 0.136 & 0.459 & 22.49 & 0.197 & 1.393 \\
CS-MAE & \underline{64.29} & \underline{0.080} & \underline{0.354} & 44.07 & 0.133 & 0.449 & \underline{23.25} & 0.197 & 1.387 \\
PIS & 47.17 & 0.183 & 0.845 & 34.65 & 0.183 & 0.569 & 19.73 & 0.217 & 1.409 \\
SMLFR & 58.68 & 0.094 & 0.475 & 35.05 & 0.177 & 0.550 & 19.47 & 0.208 & 1.562 \\
SpectralGPT & 56.00 & 0.105 & 0.494 & 39.91 & 0.159 & 0.526 & 18.85 & 0.212 & 1.464 \\
OFA-L & 60.63 & 0.090 & 0.444 & 40.46 & 0.159 & 0.519 & 20.22 & 0.206 & 1.443 \\
DyVis & 58.75 & 0.102 & 0.437 & 41.27 & 0.150 & 0.500 & 21.97 & 0.200 & 1.439 \\
\hline
Ours-B & 62.78 & 0.094 & 0.393 & \underline{45.05} & \underline{0.129} & \underline{0.439} & 23.70 & \underline{0.194} & \underline{1.372} \\
Ours-L & \textbf{65.75} & \textbf{0.077} & \textbf{0.350} & \textbf{48.12} & \textbf{0.118} & \textbf{0.411} & \textbf{24.46} & \textbf{0.182} & \textbf{1.273} \\
\hline
\end{tabular}
}
\end{table*}

\subsubsection{Evaluation on Monocular Depth Estimation Tasks}

\paragraph{Dataset Description:}
We evaluate our method on the monocular depth estimation task across three high-resolution remote sensing datasets: \textbf{Potsdam}, \textbf{Vaihingen}, and \textbf{US3D}. These datasets provide georeferenced aerial images along with accurate DSM, which are treated as ground-truth depth maps for training and evaluation.

All datasets use the \textit{NirRG} spectral band combination (near-infrared, red, green) as input, following standard practices in remote sensing analysis. To assess model robustness under limited supervision, only \textit{50\% of the original training samples are used} during fine-tuning. This setting reflects realistic scenarios where labeled data is scarce and emphasizes the importance of strong pre-training.

\paragraph{Model Architecture:}
The depth estimation framework is built upon the same \textbf{UPerNet} architecture used in semantic segmentation, adapted for regression tasks. Specifically, we replace the final classification head with a single-channel regression output that predicts per-pixel depth values. The backbone remains a pre-trained vision transformer (Ours-B or Ours-L), which extracts multi-scale features from the input image.

To handle depth regression effectively, we apply logarithmic scaling to the depth targets during training and use a hybrid loss function combining RMSE and SiLog losses, following recent monocular depth estimation literature. This ensures stable training and better metric depth recovery.

\paragraph{Evaluation Metrics:}
We adopt widely used metrics in monocular depth estimation: \textbf{D1 ($\uparrow$)}: Percentage of pixels with relative error less than 1. \textbf{RMSE ($\downarrow$)}: Root Mean Square Error between predicted and ground truth depths. \textbf{SiLog ($\downarrow$)}: Scale-invariant log error, which measures depth consistency without being affected by absolute scale differences. These metrics comprehensively reflect both accuracy and structural coherence of the estimated depth maps.

\paragraph{Results and Analysis:}
Table~\ref{tab:depth_results} summarizes the performance comparison of different pre-training strategies on all three datasets. Our proposed methods, \textbf{Ours-B} and \textbf{Ours-L}, achieve consistently superior results across all metrics.

On the \textbf{Potsdam} dataset: Ours-B achieves a D1 score of \textbf{62.78}, RMSE of \textbf{0.094}, and SiLog of \textbf{0.393}, outperforming existing self-supervised and supervised methods.  Ours-L further improves performance to \textbf{65.75}, \textbf{0.077}, and \textbf{0.350}, outperforming other methods by a significant margin.

On the \textbf{Vaihingen} dataset: Ours-L significantly surpasses methods like OFA-L and CS-MAE, achieving \textbf{48.12 D1}, \textbf{0.118 RMSE}, and \textbf{0.411 SiLog}. This indicates its stronger ability to capture fine-grained terrain variations and preserve spatial structure. On the \textbf{US3D} dataset:  Ours-L obtains the highest D1 score of \textbf{24.46} and the lowest RMSE of \textbf{0.182}, demonstrating robust generalization on large-scale urban scenes. 

Overall, these results demonstrate that our pre-training strategy enables strong transferability to depth estimation tasks, even under limited-data settings. The integration with UPerNet provides an effective framework for regressing dense depth maps from high-resolution remote sensing imagery.

\begin{figure}[!h]
    \centering
    \includegraphics[width=0.7\textwidth]{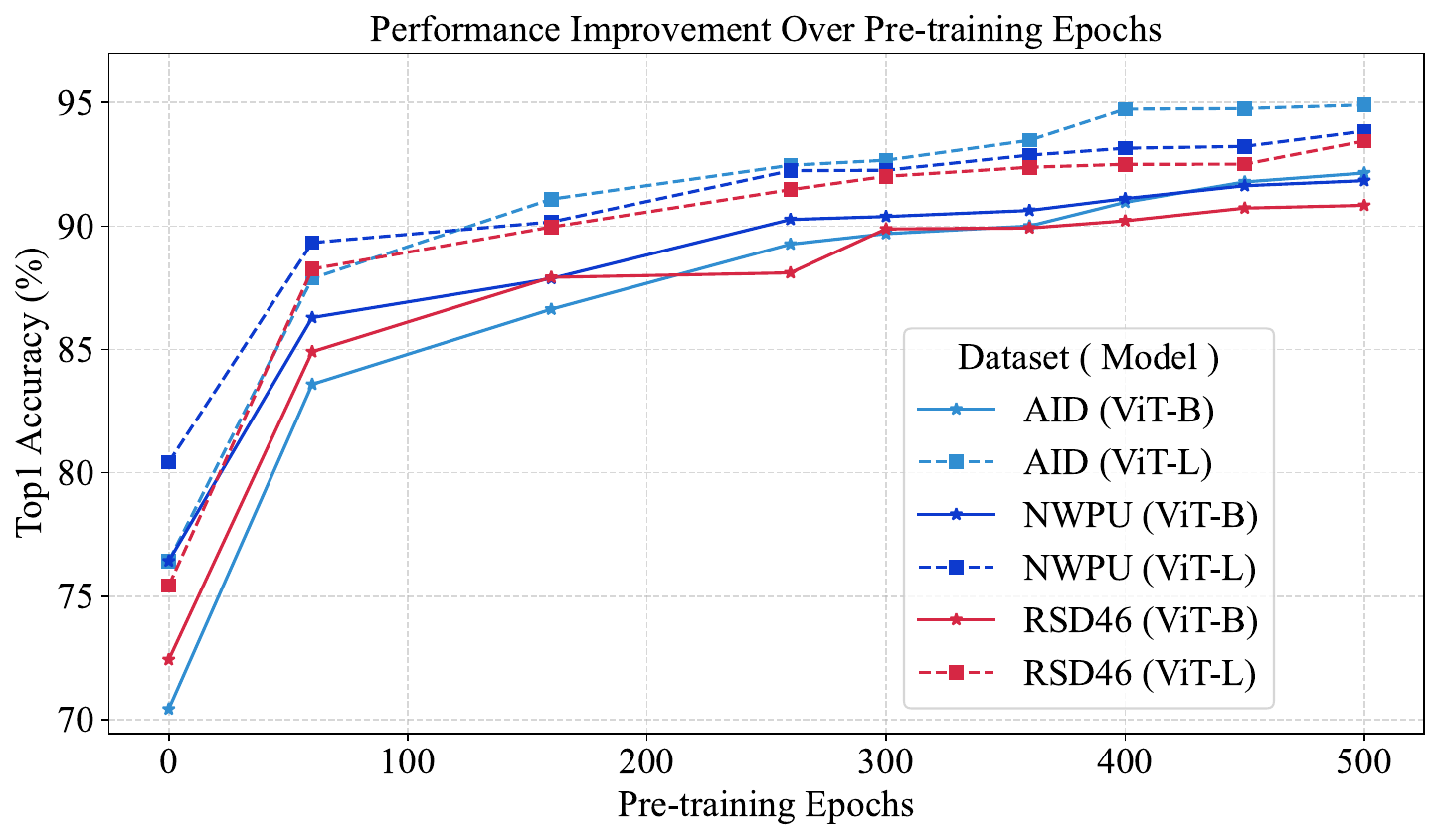}
    \caption{Performance improvement of models on three datasets (AID, NWPU, and RSD46) with different pre-training epochs.}
    \label{fig:pretrain_epochs}
\end{figure}

\subsection{Impact of Pre-training Epochs on Model Performance}

We investigate the influence of varying pre-training epochs across multiple datasets, including AID, NWPU, and RSD46. As illustrated in Figure~\ref{fig:pretrain_epochs}, the model's Top-1 accuracy improves significantly during the early stages of pre-training. 

This initial phase, particularly within the first 50 epochs, is crucial for capturing foundational semantic knowledge, which directly contributes to enhanced performance on downstream tasks. However, as the number of pre-training epochs increases beyond a certain threshold (approximately 400 epochs in our experiments), the rate of performance improvement diminishes, and the accuracy curves begin to plateau. This suggests that further extending the pre-training duration yields diminishing returns and may even lead to overfitting or unnecessary computational overhead. 

Notably, larger architectures such as ViT-L generally benefit more from extended pre-training compared to their smaller counterparts like ViT-B, especially on large-scale datasets such as NWPU. These findings underscore the importance of selecting an appropriate number of pre-training epochs based on task requirements and available computational resources to achieve an optimal trade-off between training efficiency and model performance.

\subsection{Impact of Different Backbones on Model Performance}
\begin{figure*}[h]
    \centering
    \includegraphics[width=1\textwidth]{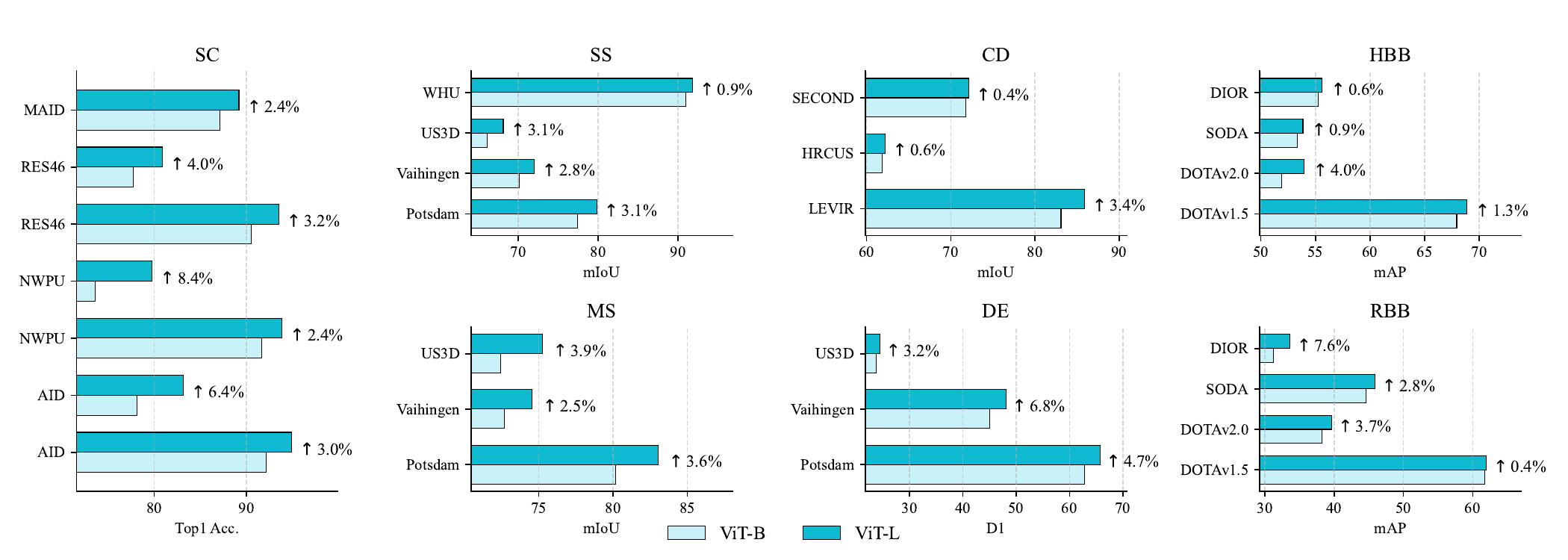}
    \caption{Impact of different model backbones. The $\uparrow$ along with the number in the figure represent the numerical improvement in accuracy of ViT-L compared to ViT-B. }
    \label{fig:backbone_impact}
\end{figure*}

We investigate the influence of different model backbones (ViT-B vs. ViT-L) on the performance of pre-trained Vision Transformer (ViT) models across various image-related tasks, including scene classification, semantic segmentation, change detection, depth estimation, and object detection. As illustrated in Figure~\ref{fig:backbone_impact}, the choice of backbone significantly impacts the model's ability to learn effective feature representations for these tasks.

The larger backbone, ViT-L, consistently outperforms ViT-B, especially on complex tasks that require capturing fine-grained details and handling large-scale datasets. For instance, in semantic segmentation, ViT-L achieves substantial improvements over ViT-B, with gains ranging from 0.9\% to 3.1\% in mIoU. Similarly, in depth estimation, ViT-L demonstrates superior performance, particularly on datasets like Vaihingen, where it improves D1 metrics by up to 6.8\%. These results highlight the importance of utilizing larger models with deeper architectures for tasks that demand high-level semantic understanding and fine-grained detail extraction.

\begin{figure*}[!h]
    \centering
    \includegraphics[width=1\linewidth]{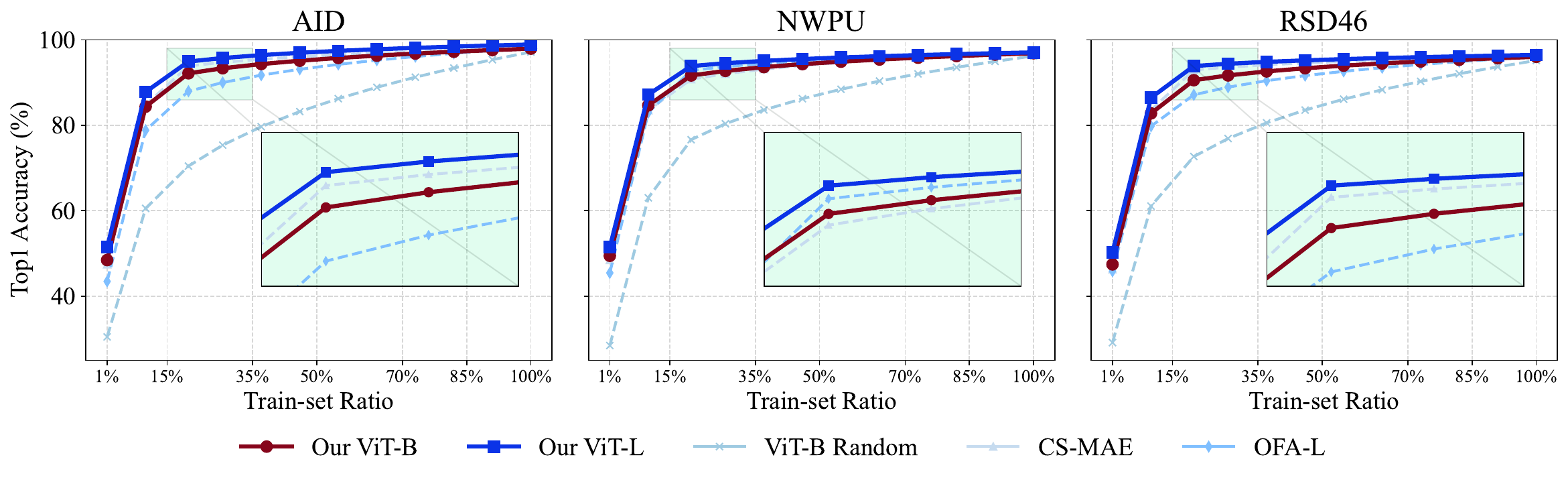}
    \caption{Impact of different train-set ratio on downstream tasks. The top-1 accuracy is plotted on the y-axis, and the train-set ratio is plotted on the x-axis. The models are evaluated on three datasets: AID, NWPU, and RSD46. The zoomed-in regions highlight the performance of different models under 15\% to 35\% conditions.}
    \label{fig:downstream_task_data_impact}
\end{figure*}
\subsection{Model Transfer Performance under Varying Data Regimes}
\label{sec:downstream_task_data_impact}

To evaluate how pre-trained models generalize to downstream tasks under different data availability conditions, we conduct a systematic analysis on three remote sensing scene classification benchmarks: AID, NWPU, and RSD46. Specifically, we vary the training set ratio from 1\% to 100\%, fine-tune the models on each subset, and report Top-1 accuracy as the evaluation metric. This experimental setup allows us to investigate both the high-data saturation behavior and low-resource adaptability of different vision models.

As shown in the Figure~\ref{fig:downstream_task_data_impact}, the Top-1 accuracy of all models across three remote sensing scene classification datasets (AID, NWPU, and RSD46) exhibits a significant upward trend as the training set ratio increases. Notably, our ViT-L model demonstrates stronger transfer capabilities in low-data scenarios (training set ratio $< 35\%$), the performance gap between models becomes particularly pronounced. As illustrated in the zoomed-in regions of the figure, our ViT-B model exhibits exceptional few-shot learning capability. It is worth noting that while ViT-L performs better in high-data regimes, its performance slightly lags behind ViT-B in extremely low-data conditions (1\%--5\%). This suggests a complex interaction between model capacity and available data scale. Such hierarchical performance behavior provides valuable insights for practical applications under resource-constrained settings.

\subsection{ Ablation Study}
\label{sec:ablation_study}

This section presents ablation studies to validate the effectiveness of the key components proposed in this paper, including the dynamic modality fusion mechanism, cross-modal compensation strategy, adaptive masking policy, and multi-modal feature alignment method. Due to the high computational cost of full-scale pre-training, all results reported in this section are based on models ViT-B and pre-trained for 100 epochs only.

\begin{table*}[ht]
\centering
\small
\caption{Performance comparison under different modality combinations during pre-training. The downstream tasks include scene classification (NWPU), single-modal semantic segmentation (Potsdam, US3D), and multi-modal semantic segmentation (Potsdam, US3D). }
\label{tab:ablation_modality_combination}
\resizebox{\textwidth}{!}{
\begin{tabular}{l|cc|ccccc}
\hline
\multirow{2}{*}{\textbf{Experiments}} & \multicolumn{2}{c|}{\textbf{Pretrain Datasets}} & \multicolumn{5}{c}{\textbf{Downstream Tasks}}  \\
\cline{2-8}
 & fMoW & HR-Pairs & \textbf{NWPU} & \textbf{Potsdam (SS)} & \textbf{Potsdam (MS)} & \textbf{US3D (SS)} & \textbf{US3D (MS)}  \\
\hline 
RGB Only (MAE)  & $ \checkmark $  &  & 83.33 & 70.20 & 73.55 & 61.37 & 66.71 \\
RGB Only (+ HR-Pairs)  & $ \checkmark $  &  $ \checkmark $ & 84.83 & 72.11 & 74.47 & 62.20  & 67.05 \\
Multi-modal (MAE)  & $ \checkmark $  &  $ \checkmark $ & 85.21 & 72.90 & 74.50 & 63.00  & 67.27 \\
RGB + MS &  $ \checkmark $ &  & 85.71 & 73.20 & 75.19 & 62.22 & 67.30 \\
RGB + DSM &  & $ \checkmark $  & 85.80 & 73.20 & 76.66 & 63.01 & 67.94\\
\hline 
Ours & $ \checkmark $   & $ \checkmark $& \textbf{87.90} & \textbf{76.50} & \textbf{78.30} & \textbf{64.11} & \textbf{68.30} \\
\hline
\end{tabular}
}
\end{table*}
\subsubsection{Impact of HR-Pairs on Pre-training}

To investigate the role of HR-Pairs in remote sensing image pre-training, we compare performance with and without HR-Pairs under the RGB-only setting. As shown in Table~\ref{tab:ablation_modality_combination}, the model pre-trained solely on the fMoW dataset achieves strong performance across multiple downstream tasks using the same network architecture (ViT-B) and training strategy. When HR-Pairs are further introduced, the model demonstrates improved robustness to complex visual variations. For instance, the classification accuracy on NWPU increases slightly from 83.33\% to 87.90\%, indicating that HR-Pairs contribute positively to feature learning.

\subsubsection{Multi-Modal Pre-Training Enhances Downstream Performance}
To evaluate the effectiveness of our multi-modal pre-training strategy, we compare our approach with baseline methods that either ignore modality-specific structures or apply standard self-supervised learning techniques such as MAE directly on multi-source data. Specifically, we examine two representative modality combinations: RGB + MS (NirRG) and RGB + DSM (digital surface model). When these multi-modal inputs are treated simply as separate images and trained using a standard MAE framework without our proposed pre-training methodology, as the Table~\ref{tab:ablation_modality_combination} shows, the performance improvements over single-modality baselines are limited—increasing from 83.33\% to 85.71\% for RGB + MS and to 85.80\% for RGB + DSM on the NWPU benchmark.

In contrast, when the same multi-modal data is trained using our proposed multi-modal pre-training method, which incorporates HR-Pairs and treats the additional modalities as extended input channels rather than distinct modalities—the performance improves significantly. Our approach achieves an overall accuracy of 87.90\% on NWPU, representing a notable gain compared to both the standard MAE-based multi-modal training and the single-modality baselines. Similar trends are observed across other downstream tasks, including semantic segmentation on Potsdam and US3D datasets under both single-modality semantic segmentation (SS) and multi-modality semantic segmentation (MS) evaluation settings. These results demonstrate that our pre-training strategy effectively leverages multi-modal information, leading to more robust and generalizable feature representations compared to conventional approaches that do not incorporate structured multi-source supervision during pre-training.

\subsubsection{Evaluation of Cross-Modal Compensation Mechanism}

\begin{table}[t]
\centering
\caption{Different masking strategies for downstream tasks in remote sensing imagery: NWPU for scene classification tasks and Potsdam (SS) for single-modality semantic segmentation tasks. }
\label{tab:ablation_cross_modal}
\begin{tabular}{l|cc}
\hline
\textbf{Masking Strategy} & \textbf{NWPU} & \textbf{Potsdam (SS)} \\
\hline
Random & 85.75 & 73.90 \\
Information-Aware & 86.01 & 75.30 \\
\hline
+ Cross-Modal (Ours) & \textbf{87.90} & \textbf{76.50} \\
\hline
\end{tabular}
\end{table}

To evaluate the effectiveness of the proposed cross-modal masking strategy in promoting cross-modal correlation learning, we conduct comparative experiments under the following settings:

- \textbf{Random Masking (Baseline)}: Traditional random masking applied independently across modalities at a fixed ratio (75\%).
    
- \textbf{Information-Aware Masking (IAM, w/o Cross-Modal)}: Our information-aware masking without cross-modal replacement.
    
- \textbf{Ours (w/ Cross-Modal Masking)}: IAM augmented with cross-modal masking and replacement to encourage inter-modal dependency learning.

All models are pre-trained under the same architecture and fine-tuned on downstream tasks such as land-use classification and scene classification. Results demonstrate that introducing the cross-modal masking mechanism significantly improves both generalization and cross-modal reasoning capabilities.

As shown in Table~\ref{tab:ablation_cross_modal}, incorporating cross-modal masking into IAM leads to a 1.2\% improvement in classification accuracy and a 1.8\% increase in mIoU.

\begin{table*}[htb]

\centering
\small
\caption{Influence of differences loss components for different down stream tasks. The downstream tasks include scene classification (NWPU), single-modal semantic segmentation (Potsdam, US3D), and multi-modal semantic segmentation (Potsdam, US3D).}
\label{tab:ablation_losses}
\resizebox{\textwidth}{!}{
\begin{tabular}{l|cccc|ccccc}
\hline
\multirow{2}{*}{ \textbf{Experiments}}  &  \multicolumn{4}{c|}{\textbf{Loss Components}} & \multicolumn{5}{c}{\textbf{Downstream Tasks}}  \\
\cline{2-10}
& $\mathcal{L}_{\text{rec}}$ & $\mathcal{L}_{\text{align}}$ & $\mathcal{L}_{\text{hsic}}$ & $\mathcal{L}_{\text{cls}}$ & \textbf{NWPU} & \textbf{Potsdam (SS)} & \textbf{Potsdam (MS)} & \textbf{US3D (SS)} & \textbf{US3D (MS)}  \\
\hline
w/o $\mathcal{L}_{\text{rec}}$ & - & + & + & + & N/A & N/A & N/A & N/A & N/A \\
w/o $\mathcal{L}_{\text{align}}$ & + & - & + & + & 86.61 & 75.20 & 77.31 & 63.20 & 67.18 \\
w/o $\mathcal{L}_{\text{hsic}}$ & + & + & - & + & 86.53 & 75.00 & 77.81 & 63.12 & 67.22 \\
w/o $\mathcal{L}_{\text{cls}}$ & + & + & + & - & 85.90 & 75.03 & 77.51 & 63.71 & 67.80 \\
\hline
Ours (Full) & + & + & + & + & \textbf{87.90} & \textbf{76.50} & \textbf{78.30} & \textbf{64.11} & \textbf{68.30}  \\
\hline
\end{tabular}
}
\raggedright
\footnotesize{\\ \textbf{Note:}  ``N/A'' indicates that the pre-training process failed to converge when $\mathcal{L}_{\text{rec}}$ was removed, making downstream task evaluation impossible.}
\end{table*}

\subsubsection{Multi-Task Self-Supervised Objective}

To assess the contribution of each loss component in our multi-task self-supervised objective, we perform ablation experiments by selectively removing or replacing the reconstruction loss, contrastive alignment loss, linear HSIC loss, and auxiliary classification loss. All models are trained under the same architecture and pre-training strategy and are fine-tuned on downstream tasks such as scene classification and semantic segmentation.

As shown in Table~\ref{tab:ablation_losses}, removing any of the individual loss components leads to a performance drop across all downstream tasks, confirming the importance of each term in learning robust and generalizable representations. Specifically, eliminating the reconstruction loss ($\mathcal{L}_{\text{rec}}$) causes training instability and failure to converge, indicating its critical role in preserving local structural information. Removing the contrastive alignment loss ($\mathcal{L}_{\text{align}}$) results in a noticeable decrease in performance, particularly on semantic segmentation benchmarks (from 76.50\% to 75.20\% mIoU on Potsdam(SS)), highlighting its contribution to cross-modal consistency. Similarly, ablating the HSIC loss ($\mathcal{L}_{\text{hsic}}$) leads to less compact and potentially redundant feature distributions, while excluding the auxiliary classification loss ($\mathcal{L}_{\text{cls}}$) weakens the discriminative capability of the learned features. These findings demonstrate that the synergy among all four loss components is essential for effective multi-modal representation learning.

\begin{figure*}[ht]
    \centering
    \includegraphics[width=0.9\linewidth]{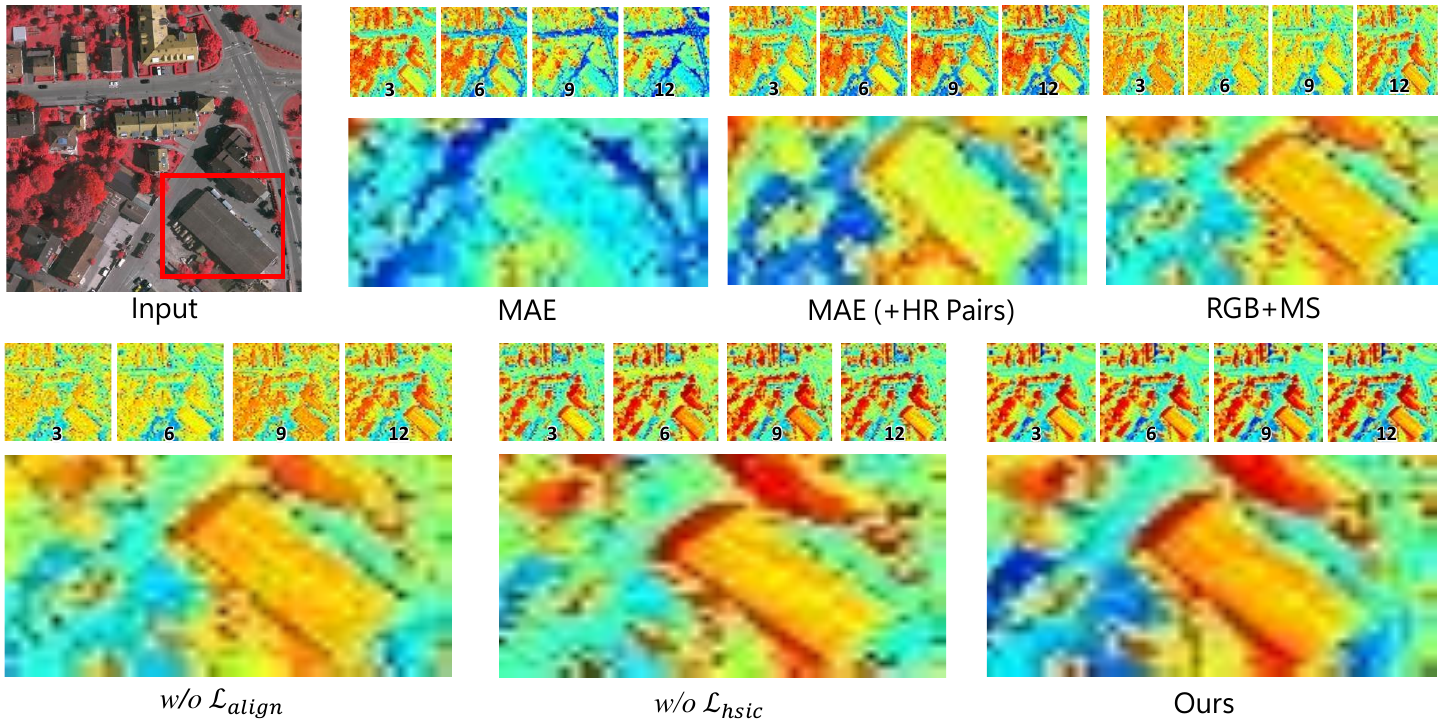}
    \caption{Comparison of feature heatmaps with different pretrain settings. The grouped small images represent the heat maps of feature maps obtained from the 3rd, 6th, 8th, and 12th layers. The larger image displays the heat map of the building areas indicated by the red boxes, which is derived from the 12th layer. Our method demonstrates the best distinction of building boundaries and various types of objects.}
    \label{fig:multi_modal_vis}
\end{figure*}

\subsubsection{Feature Map Visualization Across Pre-training Settings}

Figure~\ref{fig:multi_modal_vis} visualizes feature maps from different pre-training strategies at various network layers (layer of 3rd 6th, 9th, and 12th), providing insights into how these strategies influence feature learning. The heatmaps reveal that models pre-trained with HR-Pairs and multi-modal data (Ours) produce more structured and semantically meaningful activations compared to single-modality baselines (MAE). Notably, removing key components like the contrastive alignment loss ($\mathcal{L}_{\text{align}}$) results in weaker and less coherent feature responses, highlighting their role in learning robust cross-modal representations. Overall, the visualization complements the ablation studies by demonstrating that our full pre-training approach better captures discriminative and consistent features across layers.

\section{Conclusion}
\label{sec:conclusion}
This paper presents a novel multi-modal self-supervised distillation framework (MSSDF) and introduces a high resolution remote sensing dataset (HR-Paris), addressing the challenges of limited labeled data and complex spectral spatial heterogeneity in ultra-high-resolution (UHR) remote sensing image analysis. By integrating modality-shared feature learning, information aware adaptive masking, and multi-task self-supervised objectives, the proposed framework effectively captures both intra  and inter modal correlations, enabling robust representation learning across diverse remote sensing tasks.

Extensive experiments on 15 remote sensing datasets covering 26 downstream subtasks demonstrate the superiority of the proposed method over existing pre-training approaches such as CS-MAE, OFA-L. Ablation studies confirm the effectiveness of key components including cross-modal masking, multi-task loss design, and adaptive information preserving strategies. The HR-Paris dataset, with its centimeter-level resolution and comprehensive coverage of urban and natural landscapes, provides a valuable resource for future research in high-resolution multi-modal remote sensing analysis.

In summary, this work advances the field of remote sensing representation learning by proposing an effective self-supervised framework that leverages multi-modal fusion and knowledge distillation. Future directions include extending the framework to incorporate more sensor modalities (e.g., SAR, thermal infrared), exploring lightweight architectures for edge deployment, and applying the learned representations to real-world applications such as disaster monitoring and smart city planning.

\section*{Acknowledgement}

The numerical calculations in this paper have been done on the supercomputing system in the Supercomputing Center of Wuhan University. Tribute to open-source contributors, and gratitude to the authors of open-source works such as PyTorch, GDAL, Open-MMLab, MMSegmentation, Open-CD, and MMRotation.

\section*{Declaration of generative AI in scientific writing}
During the preparation of this work the author(s) used GPT-4o in order to grammatical modification and polish. After using this tool/service, the author(s) reviewed and edited the content as needed and take(s) full responsibility for the content of the publication.

\section*{Funding}
This work was supported by the National Natural Science Foundation of China [grant number 42101346]; the Science and Technology Commission of Shanghai Municipality [grant number 22DZ1100800]; and the China Postdoctoral Science Foundation [grant number 2020M680109].

\printcredits

 \bibliographystyle{elsarticle-num-names} 
 \bibliography{main}

\end{document}

\title{Supplementary of MSSDF: Modality-Shared Self-supervised Distillation for High-Resolution Multi-modal Remote Sensing Image Learning}

\maketitle

\begin{table}[htbp]
   \centering
   \caption{HR-Pairs Dataset Source Position Details}  \label{tab:hrinfo1}
   \begin{tabular}{c|c|c|c}
       \toprule
         & Center Coord. (Lon., Lat.)         & Country           & Area ($km^2$) \\
        \midrule
    
    1 & (51.470432, 7.050646) & Germany & 3.10 \\
        2 & (29.643957, 111.725336) & China & 19.91 \\
        3 & (30.949240, 115.141850) & China & 0.79 \\
        4 & (31.043096, 115.075789) & China & 0.59 \\
        5 & (51.470432, 7.050646) & Germany & 3.10 \\
        6 & (31.288470, 115.315613) & China & 1.15 \\
        7 & (35.674638, 139.742311) & Japan & 3.33 \\
        8 & (-41.191145, 174.912629) & New Zealand & 2.92 \\
        9 & (28.435451, -81.411610) & United States & 1.50 \\
        10 & (28.474791, -81.299681) & United States & 1.40 \\
        11 & (29.643957, 111.725336) & China & 19.91 \\
        12 & (39.103750, -84.507465) & United States & 3.48 \\
        13 & (39.766380, -86.156333) & United States & 2.37 \\
        14 & (40.664778, -73.891125) & United States & 1.67 \\
        15 & (40.709797, -73.711908) & United States & 1.76 \\
        16 & (42.355041, -71.055101) & United States & 3.41 \\
        17 & (38.570273, -109.541694) & United States & 4.18 \\
        18 & (-26.203767, 28.046721) & South Africa & 1.66 \\
        19 & (35.358294, 139.326210) & Japan & 2.32 \\
        20 & (-25.751970, 28.177492) & South Africa & 1.98 \\
\bottomrule
\end{tabular}
\end{table}
        \begin{table}[htbp]
   \centering
  
    \caption{HR-Pairs Dataset Source Position Details}   \label{tab:hrinfo2}
    \begin{tabular}{c|c|c|c}
       \toprule
         & Center Coord. (Lon., Lat.)         & Country           & Area ($km^2$) \\
        \midrule
        21 & (28.443604, -16.287232) & Spain & 2.59 \\
        22 & (28.152294, -15.423235) & Spain & 6.32 \\
        23 & (44.975375, -93.269375) & United States & 2.95 \\
        24 & (44.950218, -93.280964) & United States & 1.95 \\
        25 & (33.659878, -84.335922) & United States & 1.92 \\
        26 & (46.815505, -71.207817) & Canada & 5.62 \\
        27 & (38.035354, 23.776467) & Greece & 1.55 \\
        28 & (48.847865, 2.444652) & France & 3.15 \\
        29 & (35.507471, 139.713309) & Japan & 2.92 \\
        30 & (49.004260, 2.575696) & France & 6.85 \\
        31 & (55.914574, 23.330153) & Lithuania & 7.87 \\
        32 & (55.953712, 23.295822) & Lithuania & 5.00 \\
        33 & (59.295271, 18.024615) & Sweden & 7.31 \\
        34 & (59.339905, 18.057354) & Sweden & 6.56 \\
        35 & (64.126854, -21.819517) & Iceland & 10.69 \\
        36 & (63.842754, -22.432711) & Iceland & 9.96 \\
        37 & (-34.457056, -58.564852) & Argentina & 2.50 \\
        38 & (-34.729843, -58.339463) & Argentina & 1.91 \\
        39 & (33.319319, 130.605381) & Japan & 2.15 \\
        40 & (-38.732125, -72.584105) & Chile & 4.83 \\
            \bottomrule
\end{tabular}
\end{table}
    \begin{table}[htbp]
   \centering
   
    \caption{HR-Pairs Dataset Source Position Details}    \label{tab:hrinfo3}
    \begin{tabular}{c|c|c|c}
       \toprule
         & Center Coord. (Lon., Lat.)         & Country           & Area ($km^2$) \\
        \midrule
        41 & (-36.814446, -73.058807) & Chile & 2.62 \\
        42 & (-36.790621, -73.051292) & Chile & 0.85 \\
        43 & (-15.602985, -56.085329) & Brazil & 2.34 \\
        44 & (-3.727392, -38.525960) & Brazil & 2.13 \\
        45 & (-12.999573, -38.505664) & Brazil & 2.13 \\
        46 & (5.635202, 100.494947) & Malaysia & 1.36 \\
        47 & (-33.858104, 25.569863) & South Africa & 0.15 \\
        48 & (40.411320, -3.702693) & Spain & 1.57 \\
        49 & (7.074682, 125.602655) & Philippines & 1.47 \\
        50 & (10.337275, 123.920072) & Philippines & 1.82 \\

        51 & (-32.932462, 151.752523) & Australia & 1.63 \\
        52 & (-42.875516, 147.326317) & Australia & 2.67 \\
        53 & (-41.187066, 174.958421) & New Zealand & 4.02 \\
        54 & (31.117847, 115.105046) & China & 2.18 \\
        55 & (31.122360, 115.124355) & China & 1.59 \\
        56 & (31.043595, 115.163587) & China & 1.20 \\
        57 & (29.973663, 113.675993) & China & 0.01 \\
        58 & (30.398613, 112.689994) & China & 0.01 \\
        59 & (30.501968, 112.774032) & China & 0.01 \\
        60 & (30.420972, 112.873669) & China & 0.01 \\
                    \bottomrule
\end{tabular}
\end{table}
    \begin{table}[htbp]
   \centering
    
    \caption{HR-Pairs Dataset Source Position Details} \label{tab:hrinfo4}
    \begin{tabular}{c|c|c|c}
       \toprule
         & Center Coord. (Lon., Lat.)         & Country           & Area ($km^2$) \\
        \midrule
        61 & (29.903798, 113.594558) & China & 0.01 \\
        62 & (29.825885, 113.467134) & China & 0.02 \\
        63 & (29.807180, 113.439135) & China & 0.04 \\
        64 & (30.054721, 113.170281) & China & 4.70 \\
        65 & (30.096459, 115.759320) & China & 0.20 \\
        66 & (30.170960, 113.875906) & China & 0.02 \\
        67 & (31.032756, 121.502954) & China & 1.44 \\
        68 & (31.033434, 121.512074) & China & 1.53 \\
        69 & (31.027428, 121.503027) & China & 1.38 \\
        70 & (31.027470, 121.511906) & China & 1.29 \\
        71 & (30.460220, 114.611009) & China & 5.45 \\
        72 & (32.173942, 112.389792) & China & 36.11 \\
        73 & (30.022254, 114.912013) & China & 1.17 \\
        74 & (29.631966, 111.839783) & China & 5.89 \\
        75 & (30.525914, 114.395948) & China & 8.44 \\
        76 & (31.008366, 114.967771) & China & 0.61 \\
        77 & (29.610895, 111.710810) & China & 3.77 \\
        78 & (30.995067, 114.987511) & China & 0.45 \\
        79 & (31.030038, 114.943346) & China & 0.35 \\
        80 & (31.021997, 114.945394) & China & 0.27 \\
                    \bottomrule
\end{tabular}
\end{table}
    \begin{table}[htbp]
   \centering
   
    \caption{HR-Pairs Dataset Source Position Details}   \label{tab:hrinfo5}
    \begin{tabular}{c|c|c|c}
       \toprule
         & Center Coord. (Lon., Lat.)         & Country           & Area ($km^2$) \\
        \midrule
        81 & (31.021448, 114.982578) & China & 0.17 \\
        82 & (31.040708, 114.981497) & China & 0.30 \\
        83 & (31.048998, 114.965737) & China & 0.27 \\
        84 & (31.001381, 114.964516) & China & 2.14 \\
        85 & (31.064320, 114.966662) & China & 0.43 \\
        86 & (31.073076, 114.962093) & China & 0.21 \\
        87 & (31.074827, 114.981739) & China & 1.74 \\
        88 & (31.087440, 114.986104) & China & 2.73 \\
        89 & (31.063044, 115.004199) & China & 2.83 \\
        90 & (31.006061, 114.959206) & China & 1.18 \\
        91 & (31.051590, 114.999110) & China & 1.94 \\
        92 & (31.057828, 115.023132) & China & 1.33 \\
        93 & (31.038919, 115.057445) & China & 3.11 \\
        94 & (31.048292, 115.086146) & China & 0.48 \\
        95 & (31.008244, 114.967859) & China & 0.46 \\
        96 & (31.031306, 115.090715) & China & 1.62 \\
        97 & (31.012679, 114.966781) & China & 1.07 \\
        98 & (31.016371, 114.957204) & China & 2.17 \\
        99 & (30.995012, 114.956791) & China & 0.55 \\
        100 & (30.991485, 114.959725) & China & 0.99 \\
                    \bottomrule
\end{tabular}
\end{table}
    \begin{table}[htbp]
   \centering
  
    \caption{HR-Pairs Dataset Source Position Details}     \label{tab:hrinfo6}
    \begin{tabular}{c|c|c|c}
       \toprule
         & Center Coord. (Lon., Lat.)         & Country           & Area ($km^2$) \\
        \midrule
        101 & (31.028870, 114.958001) & China & 5.54 \\
        102 & (31.187577, 121.459388) & China & 3.13 \\
        103 & (31.210693, 115.285432) & China & 0.38 \\
        104 & (31.223095, 115.293779) & China & 1.38 \\
        105 & (31.220989, 115.281955) & China & 0.72 \\
        106 & (31.259671, 115.135977) & China & 3.58 \\
        107 & (31.253312, 115.126725) & China & 1.81 \\
        108 & (31.276543, 115.166975) & China & 1.85 \\
        109 & (31.302642, 115.212176) & China & 1.81 \\
        110 & (31.316882, 115.223938) & China & 3.11 \\
        111 & (31.331066, 115.234627) & China & 1.40 \\
        112 & (31.329486, 115.222435) & China & 1.15 \\
        113 & (31.323755, 115.187734) & China & 0.79 \\
        114 & (31.319540, 115.208474) & China & 4.03 \\
        115 & (31.339371, 115.203215) & China & 0.29 \\
        116 & (31.279973, 115.340994) & China & 4.61 \\
        117 & (31.288796, 115.355693) & China & 0.26 \\
        118 & (31.298327, 115.354703) & China & 1.09 \\
        119 & (31.357775, 115.272960) & China & 0.19 \\
        120 & (31.372223, 115.295017) & China & 2.58 \\
                    \bottomrule
\end{tabular}
\end{table}
    \begin{table}[htbp]
   \centering
    
    \caption{HR-Pairs Dataset Source Position Details}   \label{tab:hrinfo7}
    \begin{tabular}{c|c|c|c}
       \toprule
         & Center Coord. (Lon., Lat.)         & Country           & Area ($km^2$) \\
        \midrule
        121 & (31.385693, 115.357672) & China & 5.61 \\
        122 & (31.215326, 115.269987) & China & 1.25 \\
        123 & (31.247723, 115.262477) & China & 1.08 \\
        124 & (31.248021, 115.212812) & China & 0.85 \\
        125 & (31.247534, 115.168318) & China & 0.89 \\
        126 & (31.264206, 115.155452) & China & 1.29 \\
        127 & (31.006843, 115.122819) & China & 3.74 \\
        128 & (30.984663, 115.116731) & China & 1.64 \\
        129 & (30.989072, 115.125009) & China & 0.80 \\
        130 & (30.990447, 115.156941) & China & 4.46 \\
        131 & (30.967951, 115.140563) & China & 0.46 \\
        132 & (31.020943, 113.139345) & China & 7.27 \\
                \bottomrule
\end{tabular}
\end{table}

\begin{table*}[h]
\centering
\small
\caption{Quantitative results on scene classification tasks. FT(20) denotes fine-tuning with 20\% of training samples, and LP(FULL) denotes linear probing using the full training set. The best results are marked in \textbf{bold}, and the second-best are \underline{underlined}.}
\label{tab:aidcls_results}
\begin{tabular}{l|cc|cc|cc|c}
\toprule
\multirow{2}{*}{Method} & \multicolumn{2}{c|}{AID} & \multicolumn{2}{c|}{NWPU} & \multicolumn{2}{c|}{RSD46-WHU} & MAID \\
\cline{2-8}
& FT(20) & LP(FULL) & FT(20) & LP(FULL) & FT(20) & LP(FULL) & FT(FULL) \\ 
\midrule
ViT-B-Random & 70.43  & --     & 76.60   & --      & 72.66   & --      & 69.34   \\
ViT-B-ImageNet & 76.95 & 37.68  & 81.11  & 42.20    & 78.22   & 36.5    & 76.87   \\
Scale-MAE   & 92.38  & \underline{82.98}  & 91.56  & 61.88   & 90.53   & 75.96   & \underline{88.27}   \\
CS-MAE      & \underline{93.85}  & 76.57  & 90.76  & 75.22   & \underline{92.93}   & \underline{80.37}   & 85.88   \\
DMC-R       & 85.20  & 65.99  & 90.89  & 78.11   & 87.69   & 73.77   & 80.90   \\
DMC-S       & 90.67  & 82.74  & 81.18  & 63.36   & 90.29   & 76.29   & 81.30   \\
TOV         & 86.60  & 65.87  & 92.40  & 72.00   & 88.48   & 75.68   & 83.45   \\
GeRSP       & 50.60  & 22.17  & 91.82  & 71.47   & 73.26   & 47.31   & --   \\
PIS         & 92.20  & 77.42  & 92.04  & \underline{79.02}   & 92.25   & 68.94   & 87.92   \\
SMLFR       & 91.55  & 82.76  & 92.56  & 72.46   & 90.61   & 77.46   & 87.08   \\
SpectralGPT & 91.55  & 68.26  & 86.11  & 56.88   & 90.29   & 76.55   & 77.74   \\
FG-MAE-B    & 62.45  & 62.74  & 77.04  & 61.19   & 81.69   & 67.58   & 75.95   \\
FG-MAE-L    & 81.50  & 50.70  & 86.91  & 66.92   & 88.67   & 57.68   & 81.79   \\
SatMAE++    & 74.30  & 44.97  & 62.10  & 52.69   & 74.85   & 59.48   & 79.29   \\
OFA-L       & 87.95  & 71.80  & 92.82  & 75.81   & 87.11   & 61.61   & 84.38   \\
DyVis       & \underline{93.85}  & 74.76  & \textbf{93.96}  & 62.23   & 90.92   & 59.93   & 85.28   \\
\midrule
Ours-B      & 92.15  & 78.14  & 91.64  & 73.58   & 90.53   & 77.71   & 87.08   \\
Ours-L      & \textbf{94.90}  & \textbf{83.13}  & \underline{93.84}  & \textbf{79.77}   & \textbf{93.47}   & \textbf{80.85}   & \textbf{89.17}   \\
\bottomrule
\end{tabular}
\end{table*}

\begin{table*}[htb]
\centering
\small
\caption{Semantic segmentation results on Potsdam, Vaihingen, and WHU Building datasets.}
\label{tab:potsdam_seg_results_dummy}
\begin{tabular}{l|cc|cc|cc|c}
\toprule
\multirow{2}{*}{\textbf{Method}} & \multicolumn{2}{c|}{\textbf{Potsdam}} &  \multicolumn{2}{c|}{\textbf{Vaihingen}} & \multicolumn{2}{c|}{\textbf{US3D}} & \textbf{WHUB} \\
\cline{2-8}
& NirRG & NirRG-DSM & NirRG & NirRG-DSM & NirRG & NirRG-DSM & RGB \\
\midrule
ViT-B-Random & 67.10 & 67.40 & 64.44 & 65.06 & 57.92 & 63.75 & 84.78 \\
ViT-B-ImageNet & 69.65 & 70.63 & 65.34 & 66.58 & 60.36 & 66.24 & 85.66 \\
Scale-MAE & 78.35 & 80.91 & \underline{71.44} & \underline{73.06} & 67.03 & 73.79 & 90.62 \\
CS-MAE & 78.35 & \underline{81.57} & 71.37 & 71.40 & 66.44 & \underline{74.55} & 89.58 \\
DMC-R & 76.41 & 79.87 & 69.89 & 71.64 & 63.19 & 70.21 & 88.63 \\
DMC-S & 76.72 & 80.67 & 67.59 & 70.39 & 62.82 & 69.94 & 87.79 \\
TOV & 71.69 & 73.03 & 68.55 & 69.39 & 64.48 & 72.33 & 87.97 \\
GeRSP & 69.65 & 73.83 & 67.03 & 69.47 & 63.22 & 69.77 & 89.58 \\
PIS & 70.77 & 76.19 & 68.85 & 70.18 & 61.85 & 69.83 & 88.42 \\
SMLFR & 76.22 & 73.64 & 69.58 & 71.68 & 65.21 & 72.10 & 89.91 \\
SpectralGPT & 75.75 & 78.49 & 70.01 & 72.37 & 63.73 & 71.67 & 86.95 \\
FG-MAE-B & 74.61 & 76.69 & 69.19 & 70.53 & 62.42 & 68.11 & 89.45 \\
FG-MAE-L & 76.35 & 78.57 & 69.37 & 71.40 & 63.83 & 72.00 & 87.02 \\
SatMAE++ & 76.76 & 80.85 & 67.46 & 69.71 & 64.73 & 71.74 & 90.77 \\
OFA-L & \underline{78.36} & 81.42 & 70.81 & 72.38 & \underline{67.41} & 73.05 & \underline{91.59} \\
DyVis & 76.63 & 79.22 & 67.84 & 67.63 & 65.42 & 72.63 & 90.70 \\
\midrule
Ours-B & 77.42 & 80.18 & 70.11 & 72.70 & 66.08 & 72.44 & 90.94 \\
Ours-L & \textbf{79.81} & \textbf{83.03} & \textbf{72.04} & \textbf{74.54} & \textbf{68.12} & \textbf{75.24} & \textbf{91.79} \\
\bottomrule
\end{tabular}
\end{table*}

\begin{table*}[htb]
\centering
\caption{Change Detection Results on LEVIR-CD, HRCUS-CD, and SECOND. The best results are highlighted in bold. We report mIoU (\%) for each method across the datasets.}
\label{tab:cd_results}
\begin{tabular}{lcccc}
\toprule
\textbf{Method} & \textbf{LEVIRE-CD} & \textbf{HRCUS-CD} & \textbf{SECOND-BCD} & \textbf{SECOND-SCD}\\
\midrule
ViT-B-Random & 71.41 & 55.15 & 65.24 & 34.45 \\
ViT-B-ImageNet & 73.56 & 56.80 & 67.51 & 39.18 \\
Scale-MAE & 84.58 & 61.85 & 71.59 & 47.44 \\
CS-MAE & 83.06 & \textbf{62.62} & \underline{71.83} & \underline{44.47} \\
DMC-R & 75.55 & 56.92 & 70.83 & 45.79 \\
DMC-S & 80.83 & 59.84 & 68.92 & 41.29 \\
TOV & 79.57 & 58.32 & 68.34 & 39.00 \\
GeRSP & 79.09 & 60.40 & 66.34 & 35.62 \\
PIS & 78.71 & 59.15 & 68.42 & 40.96 \\
SMLFR & 80.15 & 59.20 & 68.73 & 41.66 \\
SpectralGPT & 81.28 & 60.91 & 68.88 & 41.10 \\
FG-MAE-B & 79.57 & 58.07 & 66.66 & 36.74 \\
FG-MAE-L & 82.75 & 60.47 & 68.20 & 40.15 \\
SatMAE++ & 83.83 & 61.74 & 70.18 & 44.55 \\
OFA-L & \underline{84.79} & 61.18 & 71.37 & 45.51 \\
DyVis & 82.81 & 61.68 & 70.83 & 45.55 \\
\midrule
Ours-B & 83.09 & 61.86 & 71.81 & 47.07 \\
Ours-L & \textbf{85.90} & \underline{62.21} & \textbf{72.01} & \textbf{47.51} \\
\bottomrule
\end{tabular}

\end{table*}

\begin{figure*}[htbp]
    \centering
    \includegraphics[width=\textwidth]{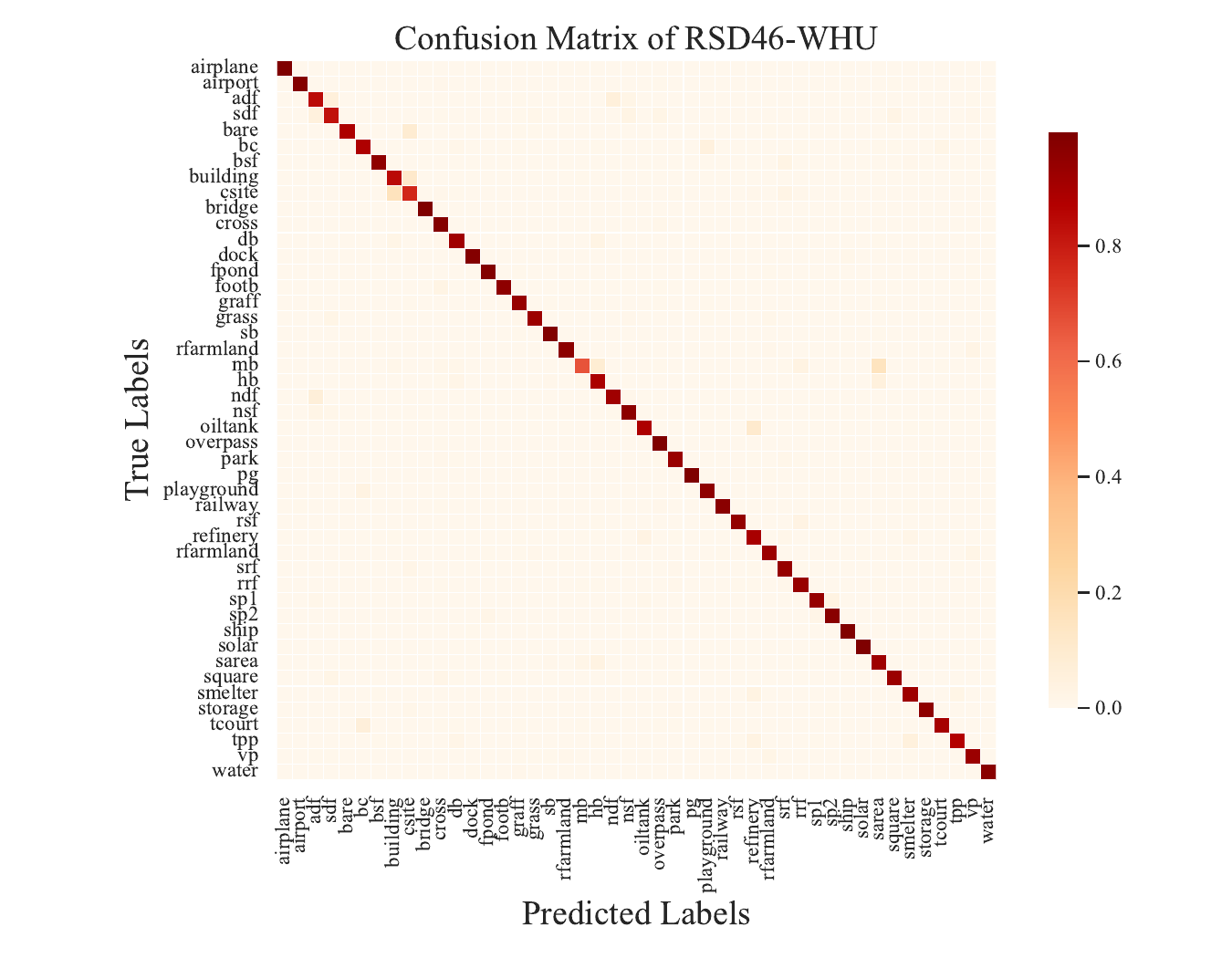}
\caption{Confusion matrix of our proposed method (Ours-L) on RSD46-WHU.}
\end{figure*}

\begin{figure*}[htpb]
    \centering
    \includegraphics[width=\textwidth]{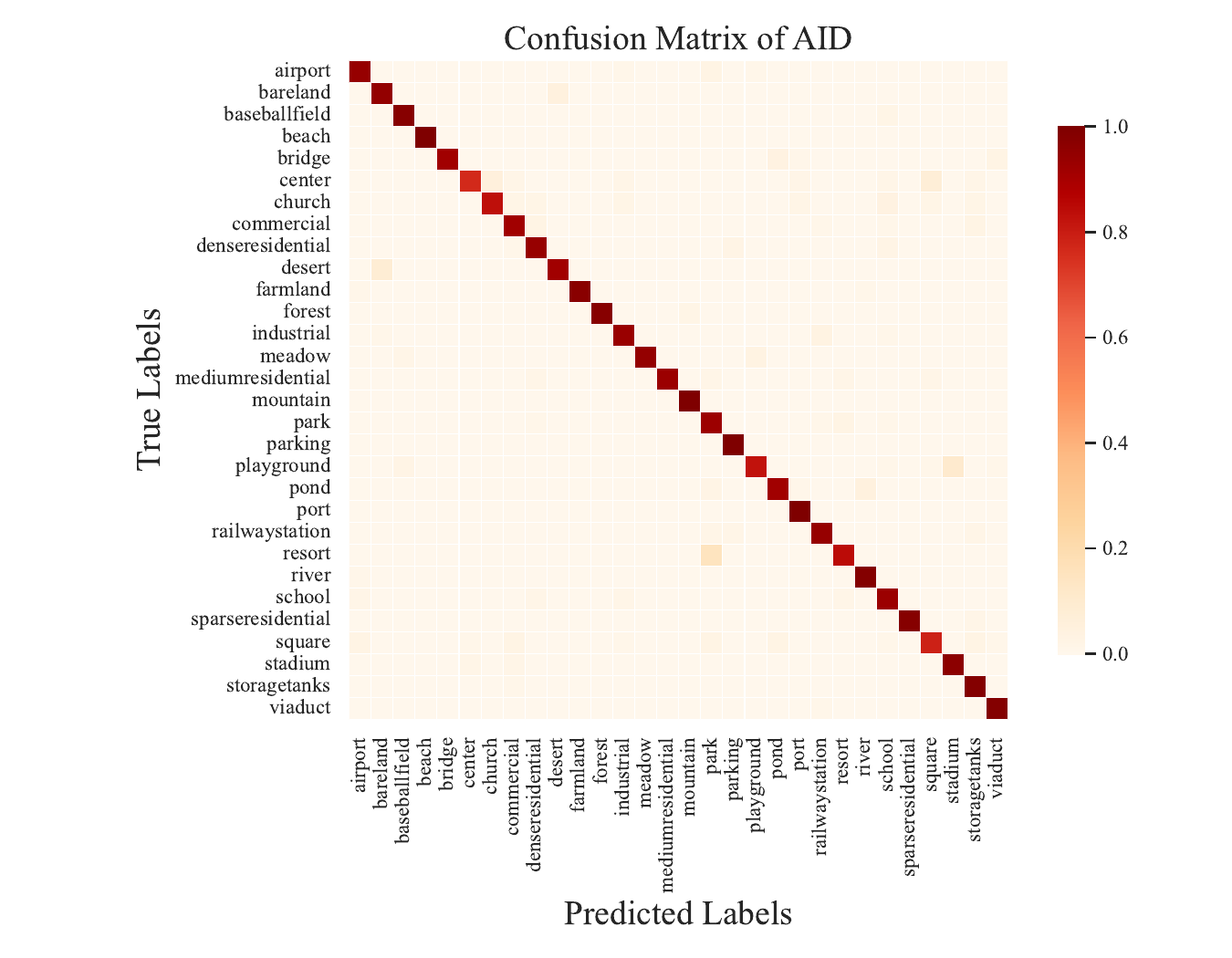}
\caption{Confusion matrix of our proposed method (Ours-L) on AID.}
\end{figure*}

\begin{figure*}[htpb]
    \centering
    \includegraphics[width=\textwidth]{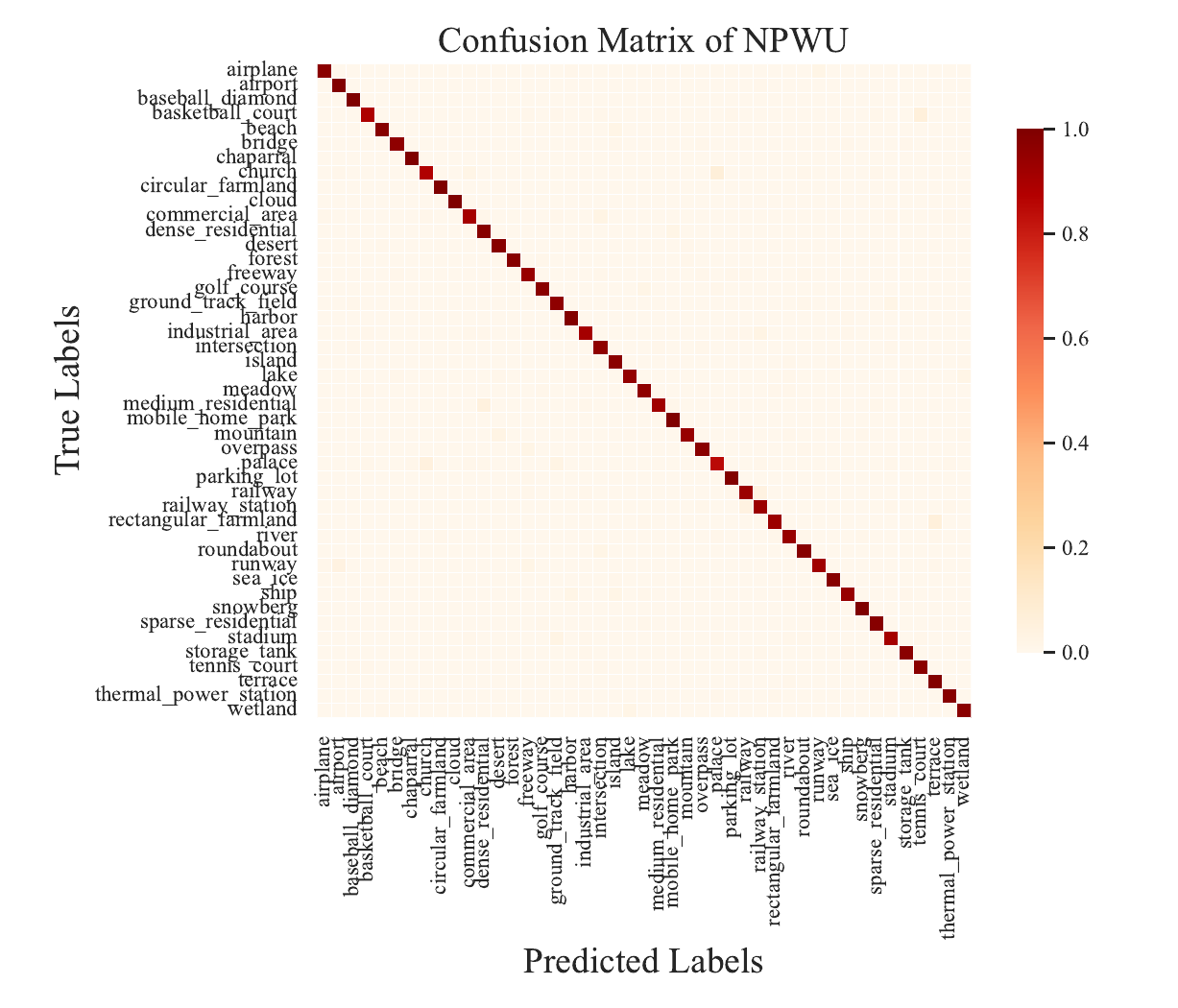}
\caption{Confusion matrix of our proposed method (Ours-L) on NWPU.}
\end{figure*}

\section{HR-Pairs Dataset Source Details}

The Table~\ref{tab:hrinfo1}, Table~\ref{tab:hrinfo2}, Table~\ref{tab:hrinfo3}, Table~\ref{tab:hrinfo4}, Table~\ref{tab:hrinfo5}, and Table~\ref{tab:hrinfo6}  presents detailed information about the original HR dataset, including latitude and longitude, area, country.

\section{Confuison Matrix of Scene Classification Tasks}
To gain deeper insights into the performance of our proposed model across different remote sensing scene classification datasets, we present the confusion matrices of our model (Ours-L) on three widely used datasets: RSD46-WHU, AID, and NWPU. These confusion matrices provide an intuitive visualization of the model's prediction accuracy for each class as well as the misclassification patterns, which helps in analyzing the model's discriminative capability across categories.

\section{Quantitative results of Scene Classification Tasks}
The Table~\ref{tab:aidcls_results} presents detailed quantitative results of our method on multiple remote sensing scene classification tasks. The table reports classification accuracies under two common settings: fine-tuning with only 20\% of labeled samples (FT(20)) and linear probing using the full training set (LP(FULL)), evaluated on the AID, NWPU, and RSD46-WHU datasets. Additionally, we report fully supervised results on the MAID dataset for comprehensive evaluation. Our method achieves either the best or second-best performance in most settings, demonstrating its strong feature representation capability and generalization power.

\section{Quantitative results of Semantic segmentation Tasks}
The Table~\ref{tab:potsdam_seg_results_dummy} present the performance of our method on semantic segmentation tasks for remote sensing imagery. Experiments are conducted on four widely used datasets: Potsdam, Vaihingen, US3D, and WHU Building. Evaluation metric is mean Intersection over Union (mIoU). The results show that our method outperforms existing state-of-the-art self-supervised pre-training approaches on most datasets, especially exhibiting stronger spatial modeling capabilities on high-resolution remote sensing images.

\section{Quantitative results of Change Detection Tasks}
The Table~\ref{tab:cd_results} reports the performance of our method on remote sensing change detection tasks. We evaluate our approach on three representative datasets: LEVIR-CD, HRCUS-CD, and SECOND. The main evaluation metric is mean Intersection over Union (mIoU). Experimental results demonstrate that our method achieves the best performance on multiple datasets, particularly showing greater robustness in handling complex background interference and multi-temporal registration challenges.

\begin{figure*}[p]
    \centering
    \includegraphics[width=1\linewidth]{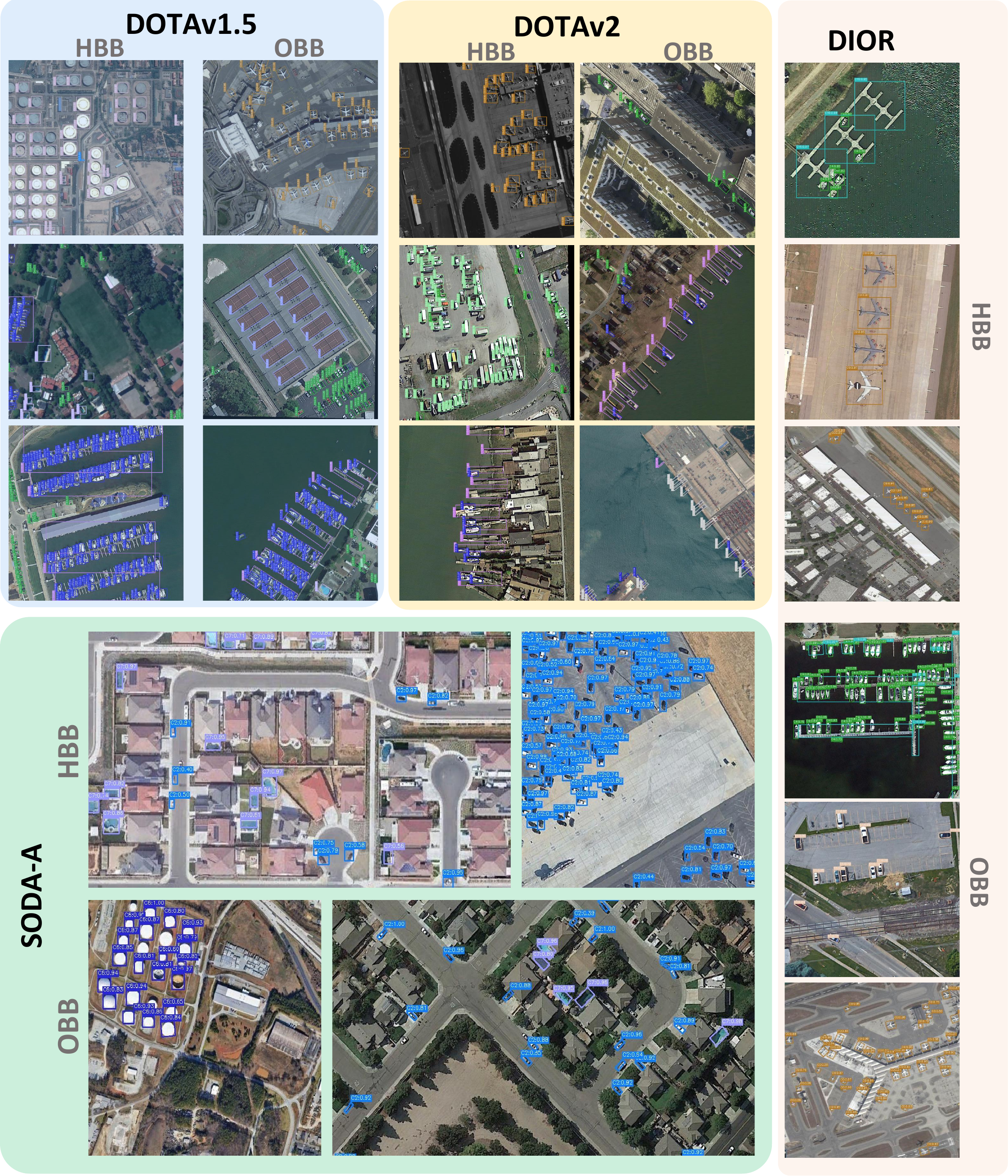}
    \caption{Qualitative results on DOTAv1.5, DOTAv2, SODA-A, and DIOR. Both the HBB and OBB tasks are include. Only the \textbf{Ours-L} model's result is include. Different colors represent different object categories.}
    \label{fig:obj_demo}
\end{figure*}

\section{Qualitative results of the Objection Detection Task }
The object detection qualitative results are shown in Figure~\ref{fig:obj_demo}. The visualizations demonstrate that our model produces  precise and tightly fitted bounding boxes. The model effectively handles occlusions, rotations, and scale variations. This further confirms the effectiveness of our pre-training strategy in improving object detection performance on remote sensing datasets.

\begin{table*}[htb]
\centering

\caption{Depth estimation results on Potsdam, Vaihingen and US3D. The best results are marked in \textbf{bold}, and the second-best are \underline{underlined}.}
\label{tab:depth_results}
\small
\resizebox{\textwidth}{!}{% 缩放到页面宽度
\begin{tabular}{l|ccc|ccc|ccc}
\toprule
\multirow{2}{*}{\textbf{Method}} & \multicolumn{3}{c|}{\textbf{Potsdam}} &  \multicolumn{3}{c|}{\textbf{Vaihingen}} & \multicolumn{3}{c}{\textbf{US3D}} \\
\cline{2-10} 
 & \textbf{D1 $\uparrow$} & \textbf{RMSE $\downarrow$ }  & \textbf{SiLog $\downarrow$}  & \textbf{D1 $\uparrow$} & \textbf{RMSE $\downarrow$ }  & \textbf{SiLog $\downarrow$}  & \textbf{D1 $\uparrow$} & \textbf{RMSE $\downarrow$ }  & \textbf{SiLog $\downarrow$}  \\
\midrule
ViT-B-Random & 38.10 & 0.167 & 0.789 & 33.43 & 0.183 & 0.564 & 14.58 & 0.250 & 1.687 \\
ViT-B-ImageNet & 39.88 & 0.160 & 0.702 & 33.59 & 0.185 & 0.573 & 14.71 & 0.244 & 1.619 \\
Scale-MAE & 59.79 & 0.090 & 0.421 & 42.50 & 0.136 & 0.459 & 22.49 & 0.197 & 1.393 \\
CS-MAE & \underline{64.29} & \underline{0.080} & \underline{0.354} & 44.07 & 0.133 & 0.449 & \underline{23.25} & 0.197 & 1.387 \\
DMC-R & 50.09 & 0.117 & 0.947 & 37.07 & 0.163 & 0.524 & 20.72 & 0.203 & 1.575 \\
DMC-S & 46.22 & 0.149 & 0.802 & 34.52 & 0.180 & 0.562 & 18.70 & 0.216 & 1.512 \\
TOV & 49.25 & 0.113 & 0.535 & 39.61 & 0.155 & 0.508 & 18.14 & 0.235 & 1.652 \\
GeRSP & 46.56 & 0.174 & 0.705 & 38.96 & 0.155 & 0.507 & 16.76 & 0.236 & 1.642 \\
PIS & 47.17 & 0.183 & 0.845 & 34.65 & 0.183 & 0.569 & 19.73 & 0.217 & 1.409 \\
SMLFR & 58.68 & 0.094 & 0.475 & 35.05 & 0.177 & 0.550 & 19.47 & 0.208 & 1.562 \\
SpectralGPT & 56.00 & 0.105 & 0.494 & 39.91 & 0.159 & 0.526 & 18.85 & 0.212 & 1.464 \\
FG-MAE-B & 50.11 & 0.195 & 0.553 & 36.75 & 0.173 & 0.544 & 19.99 & 0.220 & 1.576 \\
FG-MAE-L & 60.71 & 0.090 & 0.405 & 36.73 & 0.174 & 0.550 & 20.43 & 0.202 & 1.470 \\
SatMAE++ & 60.80 & 0.094 & 0.422 & 40.48 & 0.155 & 0.507 & 20.14 & 0.209 & 1.480 \\
OFA-L & 60.63 & 0.090 & 0.444 & 40.46 & 0.159 & 0.519 & 20.22 & 0.206 & 1.443 \\
DyVis & 58.75 & 0.102 & 0.437 & 41.27 & 0.150 & 0.500 & 21.97 & 0.200 & 1.439 \\
\midrule
Ours-B & 62.78 & 0.094 & 0.393 & \underline{45.05} & \underline{0.129} & \underline{0.439} & 23.70 & \underline{0.194} & \underline{1.372} \\
Ours-L & \textbf{65.75} & \textbf{0.077} & \textbf{0.350} & \textbf{48.12} & \textbf{0.118} & \textbf{0.411} & \textbf{24.46} & \textbf{0.182} & \textbf{1.273} \\
\bottomrule
\end{tabular}
}
\end{table*}
\section{Quantitative results of Depth estimation Tasks}
The Table~\ref{tab:depth_results} reports the performance of our model on depth estimation tasks for remote sensing imagery. We conduct experiments on three datasets that include Digital Surface Model (DSM) information: Potsdam, Vaihingen, and US3D. The evaluation metrics include the percentage of correctly estimated pixels (D1), root mean square error (RMSE), and scale-invariant log error (SiLog). The results indicate that our method significantly outperforms existing methods across all metrics, highlighting its potential in 3D remote sensing perception tasks.